\documentclass[twoside]{article}

\usepackage[accepted]{aistats2023}

\usepackage[round]{natbib}

\usepackage[utf8]{inputenc} %
\usepackage[T1]{fontenc}    %
\usepackage{hyperref}       %
\usepackage{url}            %
\usepackage{booktabs}       %
\usepackage{amsfonts}       %
\usepackage{nicefrac}       %
\usepackage{microtype}      %
\usepackage{xcolor}         %
\usepackage{wrapfig}        %

\usepackage{subcaption}
\usepackage{mathtools}
\usepackage{amsthm}
\usepackage{amssymb}
\usepackage{algorithm}
\usepackage[noend]{algorithmic}
\usepackage[pdftex]{graphicx}
\graphicspath{ {./img/} }
\usepackage{xspace}

\usepackage{placeins}

\theoremstyle{plain}

\theoremstyle{definition}

\theoremstyle{remark}

\newcommand{\name}{{\small\sf FedDrift}\xspace}

\newcommand{\mmacc}{{\small\sf FedDrift-Eager}\xspace}
\newcommand{\algfont}[1]{{\small\sf #1}\xspace}
\newcommand{\salgfont}[1]{{\scriptsize\sf #1}\xspace}
\newcommand{\ssalgfont}[1]{{\tiny\sf #1}\xspace}

\newcommand{\dist}[2]{\mathcal{P}_{#1}^{(#2)}}
\newcommand{\expec}{\mathop{\mathbb{E}}}

\DeclareMathOperator*{\argmin}{arg\,min}

\renewcommand{\cite}[1]{\citep{#1}} 

\newcommand{\McluScon}{\textbf{P1}\xspace}
\newcommand{\ScluMcon}{\textbf{P2}\xspace}

\usepackage[colorinlistoftodos,prependcaption]{todonotes}

\begin{document}

\runningauthor{Ellango Jothimurugesan, Kevin Hsieh, Jianyu Wang, Gauri Joshi, Phillip B. Gibbons}

\twocolumn[

\aistatstitle{Federated Learning under Distributed Concept Drift}

\aistatsauthor{Ellango Jothimurugesan \And Kevin Hsieh}

\aistatsaddress{Carnegie Mellon University \And Microsoft Research}

\aistatsauthor{Jianyu Wang\thanks{This work was done when Jianyu Wang was at CMU. He is now with Apple.} \And Gauri Joshi \And Phillip B. Gibbons}

\aistatsaddress{Carnegie Mellon University \And Carnegie Mellon University \And Carnegie Mellon University}

]

\begin{abstract}
Federated Learning (FL) under distributed concept drift is a largely unexplored area. Although concept drift is itself a well-studied phenomenon, it poses particular challenges for FL, because drifts arise staggered in time and space (across clients).
To the best of our knowledge, this work is the first to explicitly study data heterogeneity in both dimensions. 
We first demonstrate that prior solutions to drift adaptation that use a single global model are ill-suited to staggered drifts, necessitating multiple-model solutions. 
We identify the problem of drift adaptation as a time-varying clustering problem, and we propose two new clustering algorithms for reacting to drifts based on local drift detection and hierarchical clustering.
Empirical evaluation shows that our solutions achieve significantly higher accuracy than existing baselines, and are comparable to an idealized algorithm with oracle knowledge of the ground-truth clustering of clients to concepts at each time step.

\end{abstract}

\section{INTRODUCTION}
\label{sec:intro}

Federated learning (FL)~\cite{KonecnyMYRSB16, McMahanMRHA17} is a popular machine learning (ML) paradigm that enables collaborative training without sharing raw training data.
FL is crucial in the era of pervasive computing, where massive IoT and mobile phones continuously generate relevant ML data that cannot be easily shared due to privacy and communication constraints.
FL also enables different organizations such as hospitals~\cite{rieke2020future} and retail stores~\cite{YangLCT19} to jointly obtain valuable insights while preserving data privacy.
FL has become an important technology in the real world with massive deployments (500+ million installations on Android devices) as well as a growing market with many solution providers~\cite{flmarket}.

Existing FL solutions generally assume the training data comes from a \emph{stable} underlying distribution, and the training data in the past is sufficiently similar to the test data in the future. 
Unfortunately, this assumption is often violated in the real world, where the underlying data distribution is non-stationary and constantly evolves.
For instance, user sentiment and preference change drastically due to external environments such as the pandemic and macroeconomics~\cite{KohSMXZBHYPGLDS21, abs-2111-14938}. 
Data collected by cameras are also subject to various data changes such as unexpected weather and novel objects, which can lead to significant ML model performance losses~\cite{suprem13odin, DBLP:conf/nsdi/BhardwajXAJSKHB22, recl2023}.

This \emph{concept drift} problem
(defined in \S\ref{sec:setup}) has been studied extensively in a centralized learning environment~\cite{GamaZBPB14, TahmasbiJTG21, conf/mlsys/MallickHAJ22}. 
These centralized solutions, however, cannot address the fundamental challenges of concept drifts in FL where data is heterogeneous over \emph{time} and across different \emph{clients}. When different clients experience the data drift \emph{at different times}, no single global model can perform well for all clients. Similarly, when \emph{multiple concepts exist simultaneously}, no centralized training decision works well for all clients.
Several recent works have recognized the problem of FL under concept drift and proposed solutions that adapt learning rates or add regularization terms~\cite{chen2021asynchronous, manias2021concept, casado2021concept, guo2021towards}. 
Although these solutions perform better than drift-oblivious algorithms such as FedAvg~\cite{McMahanMRHA17}, the solutions still use a single global model for all clients, and hence fail to address the aforementioned fundamental challenges of heterogeneity over time and across clients. 
Meanwhile, centralized ensemble methods that use multiple models for adapting to drift also suffer---in response to a localized data drift, a newly created global model is trained over a mixture of concepts. The models in an ensemble are distinguished solely over time, and do not account for heterogeneity across clients.

In this work, we present the first FL solutions that employ multiple models to address FL under distributed concept drift.
Our solutions aim to create one model for each new concept so that all clients under the same concept can train that model collaboratively, similar to what is done for \emph{personalized} or \emph{clustered} FL~\citep{ghosh2019robust, ghosh2020efficient, mansour2020three, sattler2020clustered, duan2021flexible}. We introduce two new algorithms for model creation and client clustering so that our solution addresses all the challenges of distributed concept drift.
Our first algorithm, {\mmacc}, is a specialized algorithm that creates models based on drift detection. 
{\mmacc} is effective if new concepts are introduced one at a time. Our second algorithm, {\name}, is a general algorithm that leverages hierarchical clustering to adaptively determine the appropriate number of models. {\name} isolates drifted clients and conservatively merges clients via hierarchical clustering, so that {\name} can effectively handle general cases where an unknown number of new concepts emerge simultaneously.

We empirically evaluate our solution using four popular concept drift datasets, and we compare our solution against state-of-the-art centralized concept drift solutions (KUE~\citep{cano2020kappa} and DriftSurf~\citep{TahmasbiJTG21}) and a recent FL solution that adapts to concept drifts (Adaptive-FedAvg~\citep{canonaco2021adaptive}). Our results show that (i) {\mmacc} and {\name} consistently achieve much higher and more stable model accuracy than existing baselines (average accuracy 93\% vs.~90\% for the best baselines, across six dataset/drift combinations); (ii) {\name} performs much better than {\mmacc} when multiple new concepts are introduced at the same time; and (iii) our solution achieves a similar model accuracy as Oracle (94\% accuracy), an idealized algorithm that knows the timing and distribution of concept drifts. 
On the real-world drift in the FMoW dataset~\citep{KohSMXZBHYPGLDS21}, {\name} achieves 64\% accuracy vs. 58\% accuracy for the best baselines. 
We make our source code and datasets publicly available to facilitate further research on this problem.\footnote{https://github.com/microsoft/FedDrift}

\section{BACKGROUND AND MOTIVATION}

\subsection{Problem Setup}
\label{sec:setup}

We consider a FL setting with $P$ clients, assumed to be stateful and participating at each round, and a central server that coordinates training across the clients. Training data are decentralized and arriving over time. The data at each client $c = 1,\dotsc, P$ and each time $t = 1, 2, \dotsc$ are sampled from a distribution (concept) $\dist{c}{t}(x, y)$. We consider that data may be non-IID in two dimensions, varying across clients and across time. 
We say that there is a \emph{concept drift} at time $t$ and at client $c$ if $\dist{c}{t} \neq \dist{c}{t-1}$ (the standard definition of drift with respect to a single node \citep{GamaZBPB14}). Under \emph{distributed concept drift}, the time of change-points as well as the source or target distributions can differ across clients.

We seek a solution for adaptation to concept drift, generally involving any change in $\mathcal{P}(x, y)$. In contrast, by decomposing the joint distribution $\mathcal{P}(x, y) = \mathcal{P}(x)\mathcal{P}(y|x) = \mathcal{P}(y)\mathcal{P}(x|y)$, we distinguish from the special cases where $\mathcal{P}(y|x)$ is invariant (called covariate shift or virtual drift \citep{shimodaira2000improving, tsymbal2004problem, kairouz2021advances}) and $\mathcal{P}(x|y)$ is invariant (called label shift or target shift \citep{zhang13domain, azizzadenesheli19regularized}). (The datasets we consider in our evaluation (\S\ref{sec:expt}) contain general concept drifts with changes in the conditional distributions, with the exception of the FMoW dataset where the concept drift is specifically label shift.)

A single-model solution is to learn a single global model $h$ (which is a function of time but is notationally suppressed) that is used for inference at all clients. The objective is to minimize over all time $t$, 
$\sum_{c=1}^P \expec_{(x, y) \sim \dist{c}{t}}[\ell(h(x), y)]$,
where $\ell$ is the loss function. However, the optimal single model may not be well-suited in the presence of concept drifts. 
While the optimal single model can perform well under cases like covariate shift in which the feature-to-label mapping $\mathcal{P}(y|x)$ is fixed  (although achieving fast convergence still requires a specialized strategy; e.g., FedProx \citep{li2020federated}), lower loss can often be obtained under the latter case by using specialized models for different concepts.

The multiple-model option is to learn a set of global models $\{h_m\}$ for $m \in [M]$ concepts, and a time-varying clustering of clients. We denote the cluster identities by one-hot vectors $w_c^{(t)}$, where $w_{c, m}^{(t)} = 1$ when the client $c$ at time $t$ uses model $h_m$ for inference; we denote $h_{w_c^{(t)}}$ to represent the unique model $h_m$ where $w_{c, m}^{(t)} = 1$. The objective is to minimize over all time $t$, 
$\sum_{c=1}^P \expec_{(x, y) \sim \dist{c}{t}}[\ell(h_{w_c^{(t)}}(x), y)]$.

\subsection{Motivation}
\label{sec:motivation}

\begin{figure*}
\begin{minipage}{0.5\linewidth}
\centering
        \includegraphics[width=0.46\linewidth]{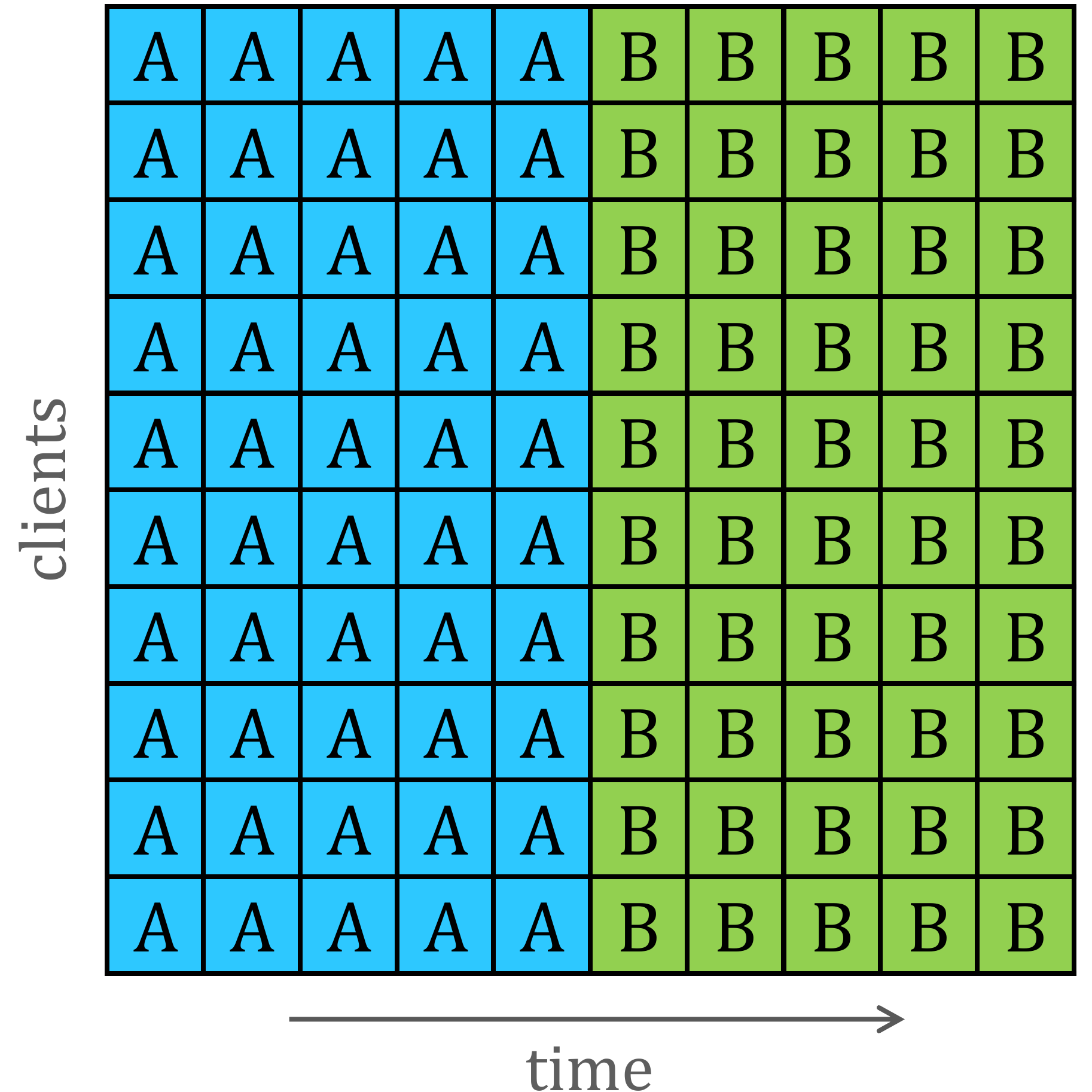}
        \includegraphics[width=0.46\linewidth]{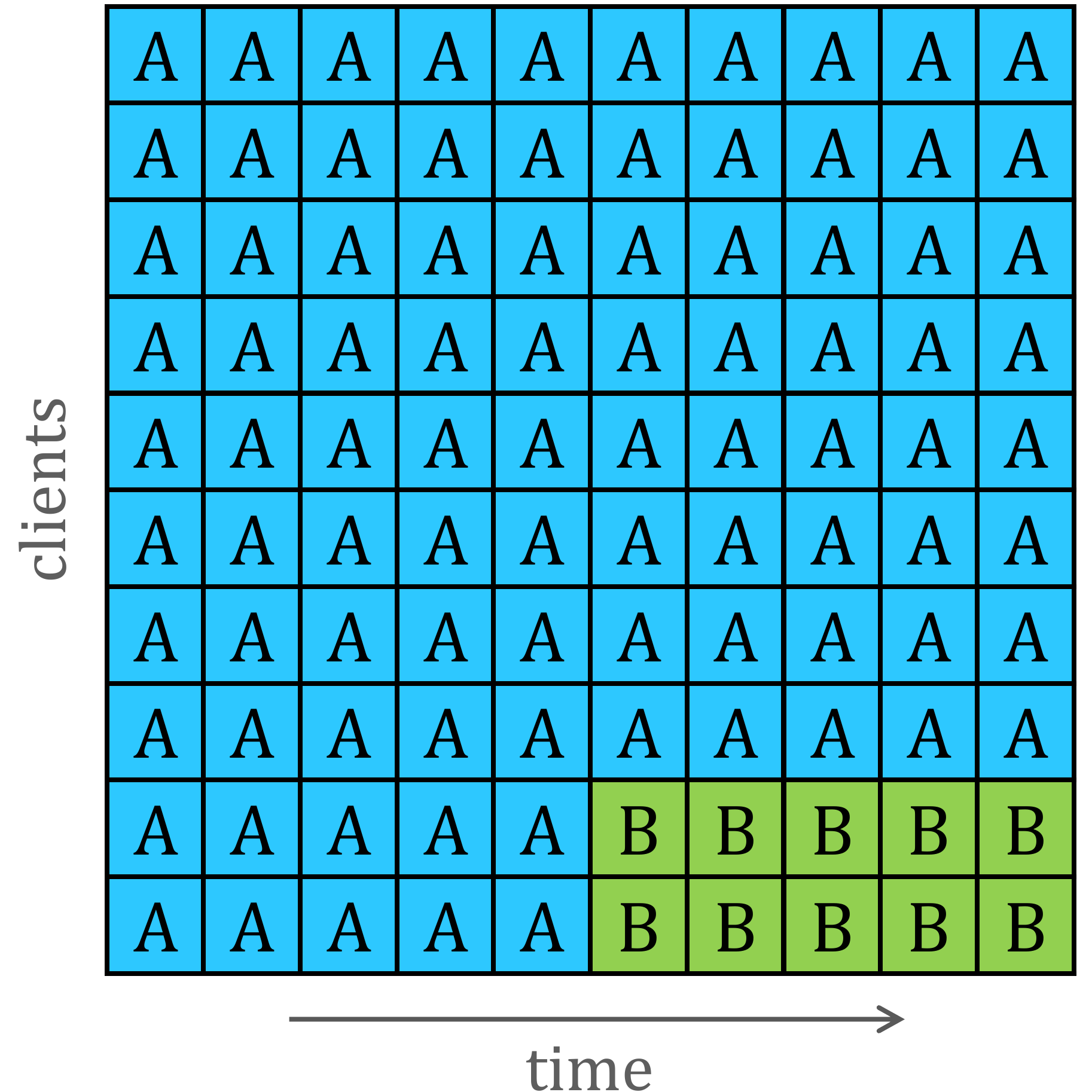}
\caption{Simplistic drifts studied in prior work. (left) Simultaneous timing. (right) One majority concept.}
\label{fig:drift-patterns-prior}
\end{minipage}%
\hfill
\begin{minipage}{0.23\linewidth}
\centering
        \includegraphics[width=\linewidth]{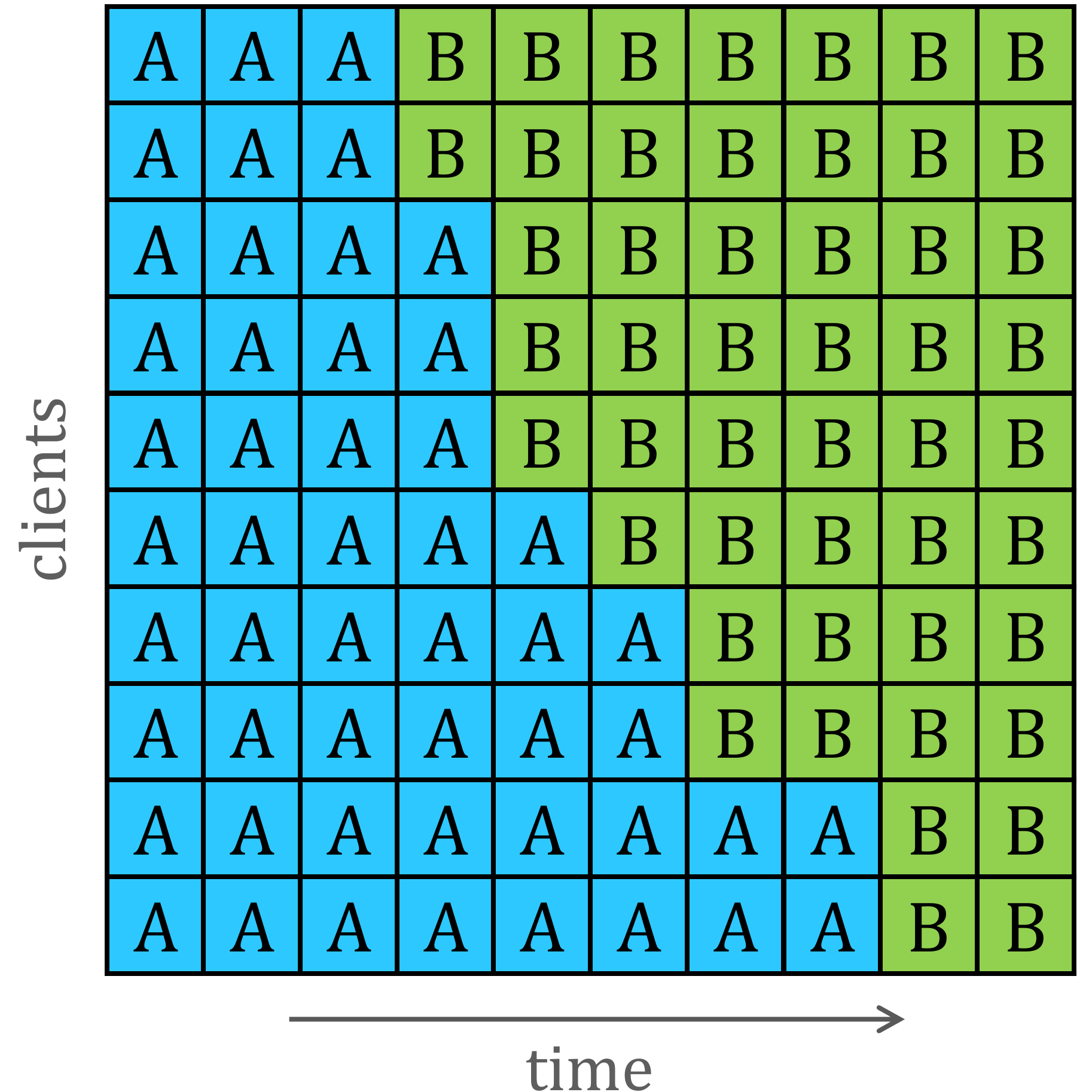}
        \caption{Distributed drift pattern (2 concepts).}
        \label{fig:2concepts}
\end{minipage}
\hfill
\begin{minipage}{0.23\linewidth}
\centering
        \includegraphics[width=\linewidth]{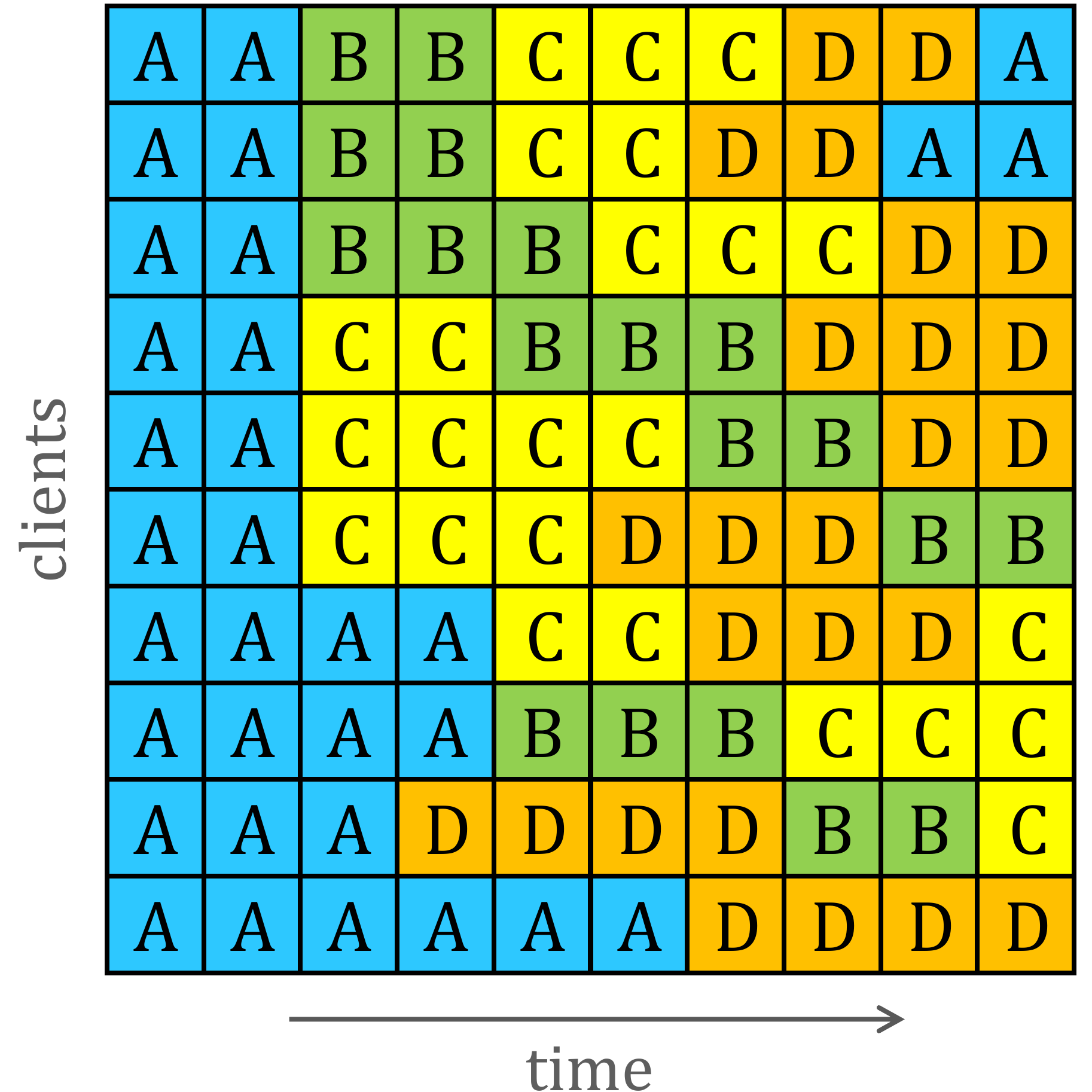}
        \caption{Distributed drift pattern (4 concepts).}
        \label{fig:4concepts}
\end{minipage}
\vspace{-0.2in}
\end{figure*}

The prior work on drift adaptation in FL only consider restrictive settings such as (i) drifts occurring simultaneously in time (e.g., Figure~\ref{fig:drift-patterns-prior}(left)), where a centralized approach works well~\citep{canonaco2021adaptive}, or (ii) drifts with only minor deviations from a majority concept (e.g., Figure~\ref{fig:drift-patterns-prior}(right)), where updates from drifting clients are suppressed and the minority concept goes unlearned~\citep{chen2021asynchronous, manias2021concept}. Our work is the first to explicitly study the more general settings arising in distributed drifts, with heterogeneous data across clients and over time. %

Consider the distributed drift pattern depicted in Figure~\ref{fig:2concepts}. This is representative of an emerging trend (e.g., a breaking news event) that effects different clients at different times (e.g., due to their lag in learning of the news). For example, consider a next word prediction app in the period when ``war'' emerges as the popular next word after ``Ukraine'' or ``slap'' emerges after ``Will Smith''. Even for this simple case of a single staggered transition between two concepts, prior work results in significant accuracy loss. In particular, their use of a single global model (and at best a single global drift detection test) results in poor accuracy during the transition period (time steps 4--8 in Figure~\ref{fig:2concepts}, see Figure~\ref{fig:time}(left) in \S\ref{sec:expt}). 

We also consider more challenging cases, as depicted in Figure \ref{fig:4concepts}, where multiple concepts emerge at the same time and concept drifts may be recurring (a.k.a.~periodic).

\begin{figure}
\centering
\includegraphics[width=0.92\linewidth]{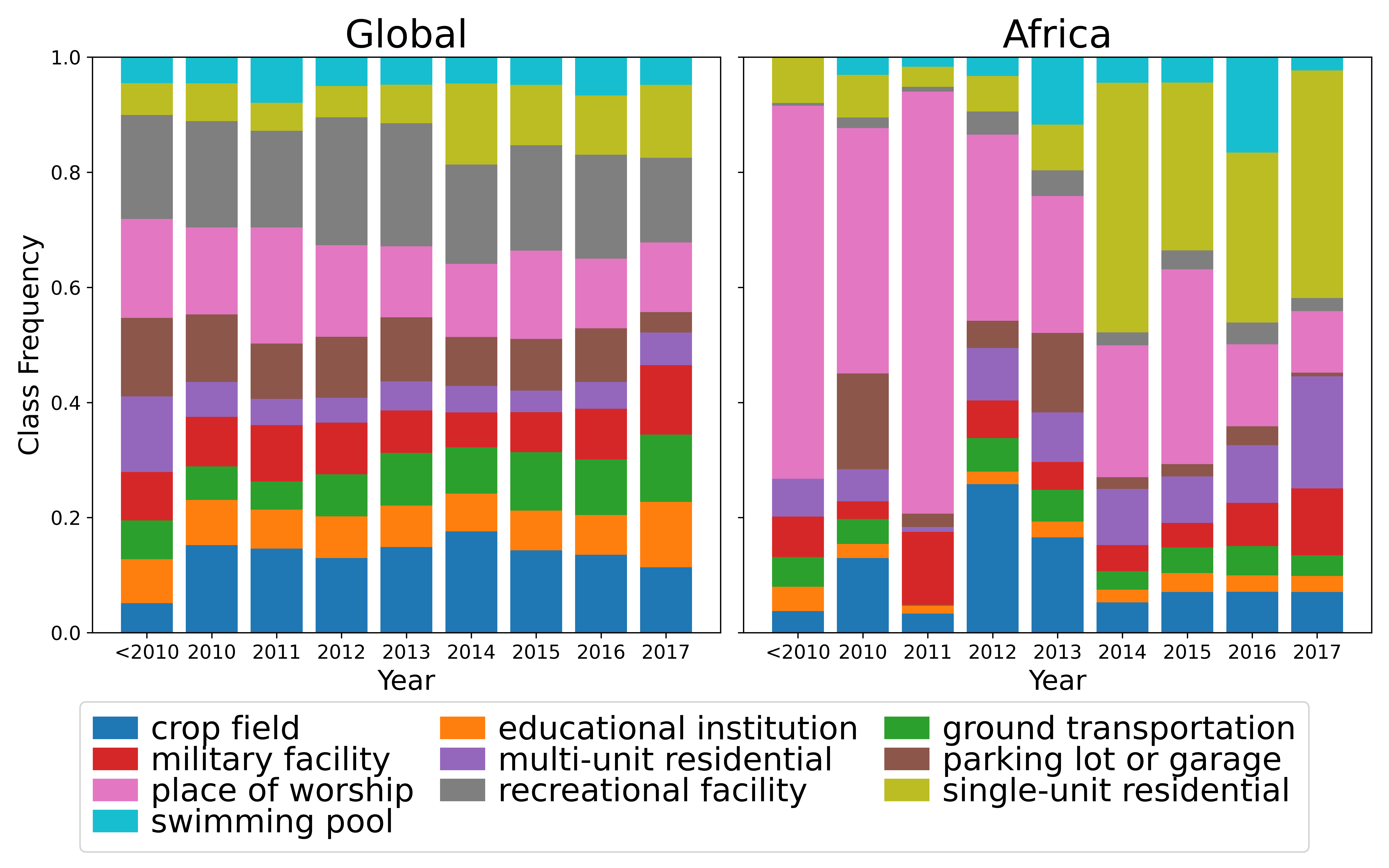}
\vspace{-0.12in}
\caption{Class distribution over time in FMoW. The data drift viewed globally (left) is small relative to the localized data drift for Africa (right).}    \label{fig:fmow-classes}
\vspace{-0.20in}
\end{figure}

To demonstrate the challenge of distributed drift in real-world data, we consider the Functional Map of the World (FMoW) dataset adapted from the WILDS benchmark \citep{christie2018functional,KohSMXZBHYPGLDS21}. The task is to classify the building type or land use from a satellite image, where images are over five major geographical regions (Africa, Americas, Asia, Europe, and Oceania) and across 16 years. Class distribution changes over time due to human activity and environmental processes. For the 10 most common classes, Figure \ref{fig:fmow-classes} shows how the class distribution in Africa changes more rapidly over time, such as a reduction in places of worship and an increase in single-unit residential buildings. However, the global class distribution is relatively slow-changing. Our evaluation shows that the model trained on the global dataset only achieves $48\%$ accuracy on Africa after the major drift at 2014, compared to $66\%$ on the rest of the world. This real-world example highlights the necessity to mitigate data drift differently across regions, and existing centralized solutions cannot address this fundamental challenge.

\vspace{-0.1in}\subsection{Related Work}
\label{sec:related}

Concept drift has been studied extensively in the centralized setting for decades. We refer the reader to the surveys by \citet{GamaZBPB14} and \citet{lu2018learning}. As previously discussed, applying these centralized algorithms to FL is not well-suited for distributed concept drifts with heterogeneous data across time and clients. We demonstrate this in our experimental evaluation (\S\ref{sec:expt}), where we compare against state-of-the-art algorithms such as KUE \citep{cano2020kappa} and DriftSurf \citep{TahmasbiJTG21}, and include concrete examples showing why their performance is worse when multiple concepts exist simultaneously.

Drift in FL, on the other hand, has so far seen only preliminary study. One line of work considers the setting where there is one concept in the system to be learned (either like the example in Figure \ref{fig:drift-patterns-prior}(right) when a minority of clients drift, or when clients observe the main concept under random noise), and seek to speed up the convergence of a model for that one concept by suppressing clients with heterogeneous data via regularization \citep{guo2021towards, chen2021asynchronous} or drift detection \citep{manias2021concept}. When it comes to adapting to a new concept over time, we are only aware of two works, and both only consider drifts with uniform timing (Figure \ref{fig:drift-patterns-prior}(left)). First, \citet{casado2021concept} consider only the \emph{covariate shift} setting (where the labeling $\mathcal{P}(y|x)$ is fixed and only $\mathcal{P}(x)$ changes) and uses drift detection to partition data from distinct concepts, in order to train a single model accurately in the course of revisiting each partition (i.e., rehearsal). Second, \citet{canonaco2021adaptive} propose Adaptive-FedAvg, in which the server tunes the learning rate used by all clients based on the variability across updates, with the goal of reacting fast when drift occurs while also achieving stable performance in the absence of drift. We compare against Adaptive-FedAvg in our evaluation. 

Our solution to drift in FL (\S\ref{sec:algo-training}, \S\ref{sec:algo-creation}) relies on learning multiple models, which has been studied in prior work on \emph{personalized FL} and \emph{clustered FL}. Clients with similar data can be grouped into clusters, where each cluster trains its global model \citep{ghosh2019robust, ghosh2020efficient, mansour2020three, sattler2020clustered, briggs2020federated, duan2021flexible}. As we extend the problem of data heterogeneity in FL with the dimension of time, we train multiple models with the algorithm in \S\ref{sec:algo-training}, which is inspired by the prior clustering algorithms IFCA \citep{ghosh2020efficient} and HypCluster \citep{mansour2020three}. This serves as the starting point of our solution, where our main contribution is the creation of new clusters as new concepts arrive over time. Our solution in \S\ref{sec:algo-creation} to handle an unknown number of concepts relies on hierarchical clustering, which has been studied in static FL by \citet{briggs2020federated}. In this prior work, it is unclear how to set the distance threshold at which to stop merging clusters. In contrast, our approach has the advantage that the stop merging criterion is identical to the drift detection threshold, which has an intuitive interpretation of performance loss.
\section{MULTIPLE-MODEL TRAINING IN FL}
\label{sec:algo-training}

\begin{table}[t]
  \centering
\caption{Table of Symbols}\vspace{-0.15in}
  \framebox{ %
  \begin{tabular}{@{}cl@{}}
    $\tau$ & current time (prior time indexed by $t$) \\
    $P$ & \# clients (indexed by $c$) \\
    $M$ & \# global models (indexed by $m$) \\
    $R$ & \# communication rounds per time (by $i$) \\
    $K$ & \# local steps per model per round (by $j$) \\
    $S_c^{(t)}$ & new data arriving at client $c$ at time $t$ \\
    $N_c^{(t)}$ & $=|S_c^{(t)}|$ \\
    $B$ & minibatch size \\
    $\eta$ & step size \\
    $h_m$ & global model $m$ \\
    $h_{c,m}$ & local update of $h_m$ by client $c$ \\
    $w_{c,m}^{(t)}$ & is $S_c^{(t)}$ used to update $h_m$? \\
  \end{tabular}}
\vspace{-0.15in}
\end{table}

As discussed above, distributed concept drift often means that multiple concepts are present simultaneously. Hence, our proposed solution learns multiple global models, where each model is trained by a cluster of clients for each distinct concept. In this section, we present Algorithm \ref{alg:training} for multiple-model training in FL for a given input clustering, which may vary over time as drifts occur. Then in \S\ref{sec:algo-creation}, we will show how to learn the necessary input clustering, and how new clusters can be created to adapt to newly appearing concepts.

We define a time step as the granularity at which new data may arrive at a client. A time step may consist of multiple communication rounds. The set of data arriving at client $c$ and time $t$ is denoted by $S_c^{(t)}$. The global models being trained are denoted by $h_m$ for $m \in [M]$, where $M$ is the total number of models at a given time. Each model is trained by a cluster of clients, where the clustering may vary over time as concept drifts occur. 
The cluster identities $w_{c,m}^{(t)}$ (\S\ref{sec:setup}) indicate whether the data $S_c^{(t)}$ that arrived at client $c$ at time $t$ are sampled when computing a local update to the global model $h_m$. Further, the cluster identity of a client at a given time indicates which model is used for inference. %

\begin{algorithm}[t]
\caption{Multiple-model training at time $\tau$}
\label{alg:training}

\begin{algorithmic}
\REQUIRE Cluster identities $w_{c,m}^{(t)}$
\FOR{each round $i = 1, 2, \dotsc, R$}
    \FOR{each client $c = 1, 2, \dotsc, P$ in parallel}
        \FOR{each model $m = 1, 2, \dotsc, M$ in parallel}
        \STATE $h_{c, m} \gets \textsc{LocalUpdate}(c, h_m, \{w_{c, m}^{(t)}\}_{t=1}^{\tau}) $
        \ENDFOR
    \ENDFOR
    \FOR{each model $m = 1, 2, \dotsc, M$}
        \STATE $h_m \gets \frac{\sum_{c=1}^P h_{c, m} \sum_{t=1}^\tau w_{c, m}^{(t)} N_c^{(t)}}{\sum_{c=1}^P \sum_{t=1}^\tau w_{c, m}^{(t)} N_c^{(t)}}$
    \ENDFOR
\ENDFOR

\STATE
\STATE
\STATE \textsc{LocalUpdate}($c, h_m, \{w_{c, m}^{(t)}\}_{t=1}^{\tau}$):
    \FOR{each local step $j=1, 2, \dotsc, K$}
        \STATE $b \gets$ random minibatch of size $B$ from $\cup_{t : w_{c, m}^{(t)} = 1} S_{c}^{(t)}$
        \STATE $h_m \gets h_m - \eta \nabla \ell(h_m; b)$
    \ENDFOR
    \STATE \textbf{return} $h_m$

\end{algorithmic}
\end{algorithm}
\begin{algorithm}[t]
\caption{Clustering to the lowest loss} %
\label{alg:clustering-loss}
\begin{algorithmic}
    \STATE $\ell_{c, m}^{(\tau)} \gets$ loss of $h_m$ on client data $S_c^{(\tau)}$
    \STATE $w_{c,m}^{(\tau)} \gets \mathbf{1}\{ m = \argmin_{m'} \ell_{c, m'}^{(\tau)} \}$
    \STATE Run Algorithm \ref{alg:training}
\end{algorithmic}
\end{algorithm}

Within each time, the training of the global models in Algorithm \ref{alg:training} is equivalent to Federated Averaging \citep{McMahanMRHA17}, since the aggregation weight of each client within each cluster is fixed at time $\tau$. So the convergence of Algorithm \ref{alg:training} can be guaranteed by directly using previous analyses for Federated Averaging, such as \cite{li2020federated, wang2021cooperative}. The difference here is that the objective function that clients are minimizing at time $\tau$ is replaced by the following:
\begin{align}
    \tilde{F}_m^{(\tau)}(h_m) = \sum_{c=1}^P \tilde{w}_{c,m}^{\tau} F_c^{(\tau)}(h_m)
\end{align}
where $F_c^{(\tau)}$ denotes the local objective function on client $c$, and the normalized weight is defined as 
\begin{align}
\tilde{w}_{c,m}^{\tau} = \frac{\sum_{t=1}^{\tau} w_{c,m}^{(t)} N_c^{(t)}}{\sum_{c=1}^P\sum_{t=1}^{\tau} w_{c,m}^{(t)} N_c^{(t)}}.
\end{align}

In the ideal case where each cluster maps to one concept in the system, each $h_m$ is specialized for each concept that is sampled from a unique data distribution ($\mathcal{P}(x,y)$), and these $h_m$ form a strong solution to our overall objective in \S\ref{sec:setup}. This ideal solution is the Oracle algorithm in our evaluation in \S\ref{sec:expt}, and we empirically demonstrate that our proposed solutions achieve comparable accuracy.

Note that, as stated, each client $c$ in Algorithm \ref{alg:training} retains its complete history of both the cluster indicators $w_{c,m}^{(t)}$ and the local data arrivals $S_{c}^{(t)}$. To reduce this overhead, each client could instead maintain just a sliding window of the most recent time steps, as long as the window suffices for the minibatch sampling in \textsc{LocalUpdate}.

Thus, we have separated the problem of concept drift in FL into two components: (i) determining the time-varying clustering of clients in response to concept drifts, which is then used as input for (ii) the multiple-model training in Algorithm \ref{alg:training}. Suppose, hypothetically, that there is a global model already initialized for each concept up to some moderate accuracy. In this restrictive setting, Algorithm \ref{alg:clustering-loss} can be used to determine the cluster identities for each new time step. Each client tests the global models from the previous time step over its newly arrived data and chooses to identify with the model with the best loss (breaking ties randomly).\footnote{If there are no new data at a particular client, then we say its cluster identity is carried over from the previous time step so the model used for inference is well-defined.} This restrictive setting covers time steps involving drifts that occur between concepts known to the system; e.g., the later stages of a staggered drift from concept A to concept B after some clients have already observed concept B (Figure~\ref{fig:2concepts}). However, Algorithm \ref{alg:clustering-loss} does not have any mechanism to spawn new clusters or determine the number of clusters. In \S\ref{sec:algo-creation}, we will show how to determine the input for Algorithm \ref{alg:training} with clustering algorithms that can spawn clusters over time to react to drifts to \emph{new} concepts.

\section{CLUSTERING ALGORITHMS}
\label{sec:algo-creation}

Under concept drift in FL, data are heterogeneous both over time and across clients. The concept at each time and client is the ground-truth clustering that we seek to learn. Ideally, the models trained by each cluster correspond 1-to-1 to the concepts present in the system. Specifically, we want to avoid two miss-clustering problems: (\McluScon) spawning \emph{multiple clusters} that correspond to a \emph{single concept}, because then each model would be trained over only a subset of the relevant data, not taking full advantage of collaborative training, and (\ScluMcon) merging clients corresponding to \emph{multiple concepts} into a \emph{single cluster} (model poisoning).

We present two clustering algorithms for adapting to concept drift. First, in \S\ref{sec:mmacc} we handle the case where only one new concept emerges at a time, which includes the example drift pattern in Figure \ref{fig:2concepts}, by incorporating a straightforward drift detection algorithm. Second, in \S\ref{sec:hierarchical} we give a general algorithm that handles the  general case where multiple new concepts may emerge simultaneously, which includes the example drift pattern in Figure \ref{fig:4concepts},  by incorporating a bottom-up technique that \emph{isolates clients} that detect drift (addressing \ScluMcon) and \emph{iteratively merges} clusters corresponding to the same concept (addressing \McluScon).

In the rest of this section, we assume that the first time step starts with one concept and one model, and that our clustering is run for each time step $\tau > 1$ as new data arrive.

\subsection{Special Case: One New Concept at a Time}
\label{sec:mmacc}

When a new concept emerges, the clients that observe the drift should be split off to a new cluster to start training a new model. Drift detection has been well-studied in the centralized, non-FL, setting \citep{gama2004learning, baena2006early, bifet2007learning, harel2014concept, pesaranghader2016fast, pesaranghader2018mcdiarmid, TahmasbiJTG21}. As we noted in \S\ref{sec:motivation}, for staggered drifts in FL, trying to apply a drift detection test \emph{globally} at the server over the aggregate error results in poor performance during the transition period. Instead, in Algorithm \ref{alg:cluster-mmacc}, we apply drift detection \emph{locally} at each client.

\begin{algorithm}[t]
\caption{\mmacc~at time $\tau$}
\label{alg:cluster-mmacc}
\begin{algorithmic}

\STATE $\ell_{c, m}^{(\tau)} \gets$ loss of model $h_m$ on client data $S_c^{(\tau)}$

\STATE $w_{c,m}^{(\tau)} \gets \mathbf{1}\{ m = \argmin_{m'} \ell_{c, m'}^{(\tau)} \}$
\IF {$\min_m \ell_{c, m}^{(\tau)} > \min_m \ell_{c, m}^{(\tau-1)} + \delta $ at any client $c$}
\STATE // \textit{create one model for all drifted clients}
    \STATE $M \gets M+1$
    \STATE Initialize a new global model $h_M$
    \STATE $w_{c,*}^{(\tau)} \gets \mathbf{0}$;  $w_{c,M}^{(\tau)} \gets 1$ 
\ENDIF

\STATE Run Algorithm \ref{alg:training}

\end{algorithmic}
\end{algorithm}

There are many drift detection tests in the literature, but the particular test is not our focus and for simplicity we consider a test of the following form.
A drift is signaled at client $c$ at time $\tau$ with respect to a model $h_m$ if the loss of the model over the newly arrived data, denoted as $\ell_{c, m}^{(\tau)}$, degrades by a threshold $\delta$ relative to the loss measured at time $\tau - 1$:
\begin{align}
\label{eq:drift-detect-one}
\ell_{c, m}^{(\tau)} > \ell_{c, m}^{(\tau-1)} + \delta.
\end{align}

This test checks for any drift that incurs performance degradation with respect to a given model. However, the desired condition for creating a \emph{new model} should check only for concept drifts that correspond to a concept \emph{previously unobserved} and \emph{ill-suited} for all existing models. For other drifts, such as the later stage of the staggered drift from concept A to concept B in Figure \ref{fig:2concepts} (after concept B has already been detected and an appropriate model created), a client should join an existing cluster (in this case, the cluster for B). Hence, in Algorithm \ref{alg:cluster-mmacc}, the drift detection test for model creation compares against the \emph{best performing} model:
\begin{align}
\label{eq:drift-detect-mult}
\min_m \ell_{c, m}^{(\tau)} > \min_m \ell_{c, m}^{(\tau-1)} + \delta.
\end{align}
\vspace{-0.1in}\\
We note that detection tests that compare across multiple models have been previously studied in centralized learning in the context of adapting to recurring drifts \citep{katakis2010tracking}.
The clustering in Algorithm \ref{alg:cluster-mmacc} (\mmacc) applies this multiple-model drift detection test at each client, and creates a new cluster for all the clients that detect a new concept; otherwise, each client identifies with the cluster with the best-performing model. This algorithm relies on the assumption that only one new concept occurs at a time by assigning the drifted clients to a single cluster. Despite this limitation, Algorithm \ref{alg:cluster-mmacc} still merits interest as it experimentally performs well on the non-trivial case of the staggered drift in Figure \ref{fig:2concepts} that has not been addressed by the prior work, as shown in \S\ref{sec:expt}. However, for the drift in Figure \ref{fig:4concepts} in which concepts B and C emerge simultaneously at different clients, this algorithm creates only one cluster and sub-optimally tries to train a single model for both new concepts (problem \ScluMcon above). Next, we extend this algorithm to address the general case where an unknown number of new concepts can occur at a time.

\subsection{General Case}
\label{sec:hierarchical}

When drifts to new concepts are detected at multiple clients, in general we do not know whether the drifts all correspond to one concept or multiple concepts (or even zero concepts in the event of false positives in detection). We designed Algorithm \ref{alg:cluster-h} (\name) for clustering in the face of this uncertainty. For each client that detects drift to a new concept, Algorithm \ref{alg:cluster-h} conservatively isolates the clients to individual clusters, and then merges clusters corresponding to the same concept slowly and safely over time by iteratively applying classical hierarchical agglomerative clustering \citep{shalev2014understanding}. 

The generic hierarchical clustering procedure is specified by a distance function over the set of elements to be clustered and a stopping criterion, and at each step until the stopping criterion is met, merges the two closest clusters, where the distance between clusters of multiple elements is commonly defined to be the maximum distance between their constituents (known as a max-linkage clustering). In Algorithm \ref{alg:cluster-h}, the {\sc Merge} subroutine combines two clusters $i$ and $j$ by averaging their models with weight proportional to the size of each model's training dataset (over all clients) and unifying the cluster identities.

To specify a distance function for hierarchical clustering, Algorithm \ref{alg:cluster-h} first aggregates at the server the loss estimates $L_{ij}$ of the model $h_i$ evaluated over a subsample of the data associated with the cluster for model $h_j$.\footnote{More precisely, at client $c$, the data clustered to $h_j$ are subsampled proportionate to the size of the local dataset relative to the global dataset for $h_j$, $\sum_t w_{c, j}^{(t)} N_c^{(t)}/\sum_{c'}\sum_t w_{c', j}^{(t)}N_{c'}^{(t)}$.} Then the distances between each cluster are initialized as $D(i, j) \gets \max(L_{ij} - L_{ii}, L_{ji} - L_{jj}, 0)$.\footnote{We note that $D(i, j)$ is not necessarily a true distance function as there is no guarantee that it satisfies the triangle inequality.} $L_{ij} - L_{ii}$ measures the loss degradation of model $h_i$ when evaluated over the data associated with $h_j$, relative to the loss over its own data. We informally interpret this difference as the magnitude of drift between the concept associated with $h_i$ to the concept associated with $h_j$, analogous to the drift detection condition in Eq (\ref{eq:drift-detect-one}) (although not identical due to the bias of $L_{ii}$ measuring a model's accuracy over its own training data). The term $D(i, j)$ is defined to be symmetric by also accounting for the magnitude of the drift $L_{ji} - L_{jj}$ in the reverse direction from concept $j$ to concept $i$.

In addition to defining the cluster distances $D(i, j)$, employing hierarchical clustering also requires setting a stopping criterion. Typically, that corresponds to specifying either the desired number of clusters (which in our case is unknown), or an upper limit on the distance between clusters to stop merging. By our identification of the cluster distance as a magnitude of drift, we naturally re-use the drift detection threshold $\delta$ to also represent the tolerance level up to which clusters can be merged, eliminating one hyperparameter.

\begin{algorithm}[t]
\caption{\name~at time $\tau$}
\label{alg:cluster-h}

\begin{algorithmic}
\STATE $\ell_{c, m}^{(\tau)} \gets$ loss of model $h_m$ on client data $S_c^{(\tau)}$

\FOR{each client $c = 1, 2, \dotsc, P$ in parallel}
\IF {$\min_m \ell_{c, m}^{(\tau)} > \min_m \ell_{c, m}^{(\tau-1)} + \delta $}
    \STATE Initialize a local model at client $c$ to be added to the set of global models at $\tau+1$, and assign client $c$ to its own cluster
\ELSE
    \STATE $w_{c,m}^{(\tau)} \gets \mathbf{1} \{ m = \argmin_{m'} \ell_{c, m'}^{(\tau)} \}$
\ENDIF 
\ENDFOR

\FOR {each $i,j$ from $1, 2, \dotsc, M$ in parallel}
\STATE $L_{ij} \gets$ loss of model $h_i$ on sample of $\cup_{c,t : w_{c, j}^{(t)} = 1} S_{c}^{(t)}$
\ENDFOR

\STATE Cluster distances $D(i,j) \gets \max(L_{ij} - L_{ii}, L_{ji} - L_{jj}, 0)$

\WHILE {$\min_{i \neq j}D(i,j) < \delta$}
    \STATE \textsc{Merge}($i, j, D$)
\ENDWHILE

\STATE Run Algorithm \ref{alg:training}

\STATE
\STATE \textsc{Merge}($i, j, D$):
    \STATE Add a new model $h_k \gets \frac{h_i \sum_{c,t} w_{c,i}^{(t)} N_c^{(t)} + h_j \sum_{c,t} w_{c,j}^{(t)} N_c^{(t)} }{ \sum_{c,t} w_{c,i}^{(t)} N_c^{(t)} + \sum_{c,t} w_{c,j}^{(t)} N_c^{(t)}}$
    \STATE $w_{c, k}^{(t)} \gets w_{c, i}^{(t)} + w_{c, j}^{(t)} $ for all $c,t$
    \STATE $D(k,l) = \max(D(i,l), D(j,l))$ for all $l$
    \STATE Delete models $h_i, h_j$

\end{algorithmic}

\end{algorithm}

In Algorithm \ref{alg:cluster-h}, both creating new clusters and merging existing clusters are based on the observed difference of the models' accuracy across two samples of data. For the clustering to accurately distinguish concepts, we assume that relevant changes in the concepts are manifested in the degradation of a model's predictive accuracy, and that the local sample size is sufficient for statistical significance---the same assumptions necessary for prior drift detection tests \citep{harel2014concept, pesaranghader2016fast, pesaranghader2018mcdiarmid, TahmasbiJTG21}.

One subtlety to Algorithm \ref{alg:cluster-h} is that the hierarchical clustering is iteratively run at every time step, because the cluster distances vary with time. A simpler alternative would be to only try merging newly created clusters of local models after one time step of training. However, at that one time step, even models corresponding to the same concept may fail to merge given the limited sample size and limited number of training iterations. In other words, while the models are still warming-up, they may still be separated by a distance exceeding $\delta$. As the models converge over time, the distance may drop below $\delta$, which Algorithm \ref{alg:cluster-h} accounts for by iteratively attempting to merge.

The hierarchical clustering strategy of Algorithm \ref{alg:cluster-h} allows it to adaptively determine the appropriate number of clusters even when an unknown number of new concepts emerge at a time, but it also incurs additional computational resources relative to Algorithm \ref{alg:cluster-mmacc}. Algorithm \ref{alg:cluster-h} creates more global models $M$, adding to the communication cost of sending $O(MP)$ models. Additionally, the hierarchical clustering adds an $O(M^2 \log M)$ time complexity at the server at every time step (using a heap data structure for finding the minimum pairwise distance). In Appendix \ref{sec:appendix-expt-results}, we discuss how we can restrict Algorithm \ref{alg:cluster-h} to create fewer overall models for higher efficiency.
Also, similar to Algorithm \ref{alg:training}, each client $c$ could maintain $w_{c,m}^{(t)}$ and $S_{c}^{(t)}$ for just a sliding window of the most recent time steps, as long as the window suffices for Algorithm \ref{alg:cluster-h}'s subsampling step.

\section{EXPERIMENTAL RESULTS}
\label{sec:expt}

We empirically demonstrate that \mmacc and \name are more effective than prior centralized drift adaptation and achieve high accuracy that is comparable to an oracle algorithm in the presence of distributed concept drifts. Prior work on FL under drifts is limited to simple cases such as in Figure \ref{fig:drift-patterns-prior}, as noted in \S\ref{sec:motivation}. Our evaluation covers the synthetic drifts in Figures \ref{fig:2concepts} and \ref{fig:4concepts}, which represent more complex scenarios where drifts (i) occur across clients with staggered timing, (ii) correspond to different concept changes across different clients, and (iii) involve recurring concepts (e.g., the sequence A--B--C--D--A). We also evaluate on the real-world drift in the FMoW dataset (\S\ref{sec:motivation}), which shows gradual concept changes staggered across clients.

The synthetic drift patterns are studied with respect to the following datasets: SINE \citep{pesaranghader2016framework}, CIRCLE \citep{pesaranghader2016framework}, SEA \citep{bifet2010moa}, and MNIST \citep{lecun1998gradient}. SINE and CIRCLE each have 2 defined concepts, and we generate partitions of the data under the 2-concept staggered drift of Figure \ref{fig:2concepts}, while SEA and MNIST have more defined concepts, and we generate partitions under both the 2-concept and 4-concept drift patterns of Figures \ref{fig:2concepts} and \ref{fig:4concepts} for 10 clients and 10 time steps. 
For the real drift in FMoW, we evaluate on a subset of the data including the 10 most common classes, and identify each of the 5 major regions as one client and each new year as one time step.
Appendix \ref{sec:appendix-expt-setup} has further dataset details.

We compare our algorithms \mmacc and \name against the following baselines. First, the \algfont{Oblivious} algorithm learns a single model with FedAvg and has no mechanism for drift adaptation. Second, we consider traditional (non-FL) drift adaptation algorithms applied centrally at the server on top of FedAvg. Drift adaptation is typically classified into three categories, and we compare against algorithms representative of each: the drift detection method \algfont{DriftSurf} \citep{TahmasbiJTG21}, 
two ensemble methods \algfont{KUE} \citep{cano2020kappa} and \algfont{AUE} \citep{brzezinski2013reacting}\footnote{By comparing against ensemble methods, we also account for the factor that multiple-model algorithms have higher capacity than single-model algorithms. \salgfont{AUE} and \salgfont{KUE} make predictions using a weighted vote over 5 and 10 models, respectively.}, and a \algfont{Window} method that forgets data older than one time step (more are reported in Appendix \ref{sec:appendix-expt-results}). Third, \algfont{Adaptive-FedAvg} \citep{canonaco2021adaptive} is an FL algorithm that learns a single model and adapts to drifts by centrally tuning the learning rate used by all clients as a function of the variability across updates. Fourth, we compare to static FL clustering algorithms \algfont{IFCA} \citep{ghosh2020efficient} and \algfont{CFL} \citep{sattler2020clustered}, which we extend to the time-varying setting by adding a window method (more variations reported in Appendix \ref{sec:appendix-expt-results}). Fifth, \algfont{Oracle} is an idealized algorithm that has oracle access to the concept ID at training time and runs the multiple-model training of Algorithm \ref{alg:training} with the ground-truth clustering. 

We run our experiments using the FedML framework \citep{chaoyanghe2020fedml}. At each time step, each client observes a new batch of training data. For all the experiments on synthetic datasets, the models trained under each algorithm are fully connected neural networks with a single hidden layer of size $2d$ where $d$ is the number of features. On the FMoW dataset, each algorithm trains ResNet18 models pretrained on ImageNet \citep{he2016deep}. After training for each time step, we test each algorithm over the batch of data arriving at the following time step, for all time steps. Each experiment is run for 5 trials, and we report the mean and the standard deviation. Additional algorithm details are in Appendix \ref{sec:appendix-expt-setup}. 

\begin{table*}[t]
\vspace{-0.10in}
\setlength{\tabcolsep}{4pt}
\caption{Average accuracy (\%) across all clients and time (over 5 trials)}
\vspace{-0.15in}
\label{table:accuracy}
\begin{center}
\fontsize{8pt}{8pt}\selectfont
\begin{tabular}{lccccccc}
\toprule
 & SINE-2 & CIRCLE-2 & SEA-2 & MNIST-2 & SEA-4 & MNIST-4 & FMoW \\
\midrule
\salgfont{Oblivious} &	52.11	$\pm$	1.79	&	88.38	$\pm$	0.17	&	86.46	$\pm$	0.22	&	87.37	$\pm$	0.16	&	85.40	$\pm$	0.09	&	82.95	$\pm$	0.03	&	58.57	$\pm$	0.07	\\
\salgfont{DriftSurf} &	84.18	$\pm$	1.40	&	92.34	$\pm$	0.38	&	87.20	$\pm$	0.27	&	93.26	$\pm$	0.52	&	85.55	$\pm$	0.13	&	82.97	$\pm$	0.09	&	58.45	$\pm$	0.19	\\
\salgfont{KUE} &	86.86	$\pm$	0.17	&	93.71	$\pm$	0.14	&	87.25	$\pm$	0.94	&	90.44	$\pm$	0.44	&	85.09	$\pm$	0.86	&	79.89	$\pm$	0.26	&	33.11	$\pm$	6.09	\\
\salgfont{AUE} &	86.00	$\pm$	0.95	&	92.84	$\pm$	0.19	&	87.48	$\pm$	0.07	&	92.22	$\pm$	0.05	&	85.47	$\pm$	0.12	&	82.07	$\pm$	0.47	&	54.23	$\pm$	0.14	\\
\salgfont{Window} &	86.28	$\pm$	0.64	&	93.72	$\pm$	0.14	&	87.94	$\pm$	0.10	&	92.34	$\pm$	0.07	&	85.72	$\pm$	0.13	&	81.43	$\pm$	0.44	&	58.88	$\pm$	0.15	\\
\ssalgfont{Adaptive-FedAvg} &	74.10	$\pm$	10.03	&	86.26	$\pm$	0.00	&	86.77	$\pm$	0.53	&	92.18	$\pm$	0.05	&	85.25	$\pm$	0.27	&	81.64	$\pm$	0.04	&	52.82	$\pm$	0.21	\\
\salgfont{IFCA+Window} &	\textbf{98.49	$\pm$	0.13}	&	94.31	$\pm$	1.62	&	\textbf{88.04	$\pm$	0.17}	&	91.76	$\pm$	0.50	&	86.17	$\pm$	1.00	&	81.27	$\pm$	0.43	&	49.40	$\pm$	0.76	\\
\salgfont{CFL+Window} &	96.92	$\pm$	1.84	&	96.04	$\pm$	1.56	&	87.81	$\pm$	0.32	&	90.66	$\pm$	0.35	&	86.06	$\pm$	0.11	&	80.51	$\pm$	0.72	&	58.82	$\pm$	0.11	\\
\midrule																												
\salgfont{FedDrift-Eager} &	97.53	$\pm$	0.13	&	\textbf{97.82	$\pm$	0.17}	&	87.51	$\pm$	0.88	&	\textbf{95.52	$\pm$	0.11}	&	87.61	$\pm$	1.26	&	90.69	$\pm$	1.20	&	61.77	$\pm$	0.51	\\
\salgfont{FedDrift} &	97.43	$\pm$	0.06	&	\textbf{97.82	$\pm$	0.19}	&	87.29	$\pm$	0.75	&	95.48	$\pm$	0.08	&	\textbf{88.13	$\pm$	0.76}	&	\textbf{93.80	$\pm$	0.08}	&	\textbf{64.84	$\pm$	0.33}	\\
\midrule																												
\salgfont{Oracle} &	98.45	$\pm$	0.03	&	97.84	$\pm$	0.22	&	87.76	$\pm$	0.98	&	95.54	$\pm$	0.11	&	88.79	$\pm$	0.41	&	94.30	$\pm$	0.08	&	-			\\
\bottomrule
\end{tabular}
\end{center}
\vspace{-0.2in}
\end{table*}

In Table \ref{table:accuracy}, we report the test accuracy averaged across all clients and all time steps except for the times of drifts (for synthetic datasets). We omit the times of drift because there is no chance for a client to adapt to the drift yet, and we eliminate the noise from beneficial clustering mistakes if by chance a client were clustered to the model appropriate for the test data after the drift. (For completeness, Appendix \ref{sec:appendix-expt-results} shows results averaged over all time steps including drifts.)

\begin{figure}
    \centering
    \includegraphics[trim=1cm 9cm 0.5cm 7.5cm, width=1.0\linewidth]{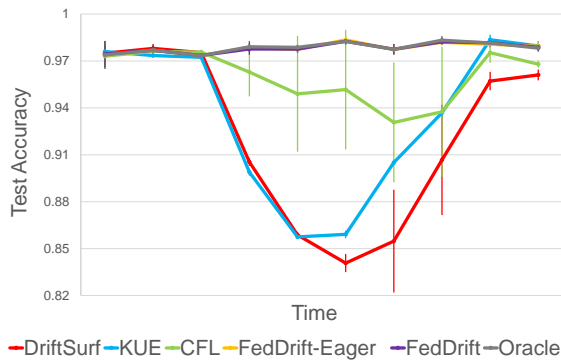}
\caption{Accuracy at each time (averaged across clients) on CIRCLE-2.}%
\label{fig:time}
\vspace{-0.25in}
\end{figure}

Across all the 2-concept datasets under the staggered drift, we observe that the multiple-model algorithms \mmacc and \name outperform the prior centralized solutions. In Figure \ref{fig:time}, the accuracy is broken down per time step on CIRCLE-2, where we observe that centralized algorithms particularly suffer during the transition period. The fundamental issue is that when both concepts simultaneously exist, no single model can accurately fit for all clients. Even the ensemble algorithm (\algfont{KUE}) has poor performance because any new model added is updated by each client, and during the transition period, there is no model trained solely over data from the second concept. \mmacc and \name learn models specialized for the second concept immediately after it emerges, and learn to apply the appropriate model at each client during the transition, matching the performance of \algfont{Oracle}.

Another challenge that the 2-concept staggered drift poses for \algfont{DriftSurf}, \algfont{KUE}, \algfont{AUE}, and \algfont{Adaptive-FedAvg} is that their adaptation strategies are a function of estimators that, from the central server's perspective, are aggregated over some clients that are drifting and others that are not. It is muddy whether drift is truly occurring, and even the unsophisticated window-based algorithm performs slightly better.

The clustering algorithms \algfont{IFCA} and \algfont{CFL} with a window perform relatively well on the 2-concept staggered drifts because they can flexibly employ a model specialized for the second concept during the transition period, but are overall behind \name and \mmacc. We observe \algfont{IFCA}'s success in adapting to drift is dependent on its random parameter initialization for its clusters, and works well particularly for the sharp drift on SINE-2.\footnote{The accuracy of \salgfont{IFCA} is higher than \salgfont{Oracle} in a few cases but within the standard deviation, which we attribute to randomness in the model initialization and training.} For \algfont{CFL}, we observe that its iterative cluster splitting reacts quickly to drift, but creates excessive models for a concept over time without unifying clients under staggered drift. Appendix \ref{sec:appendix-expt-results} has more details.

\begin{wrapfigure}{R}{0.5\linewidth}
\vspace{-0.15in}
        \includegraphics[width=\linewidth]{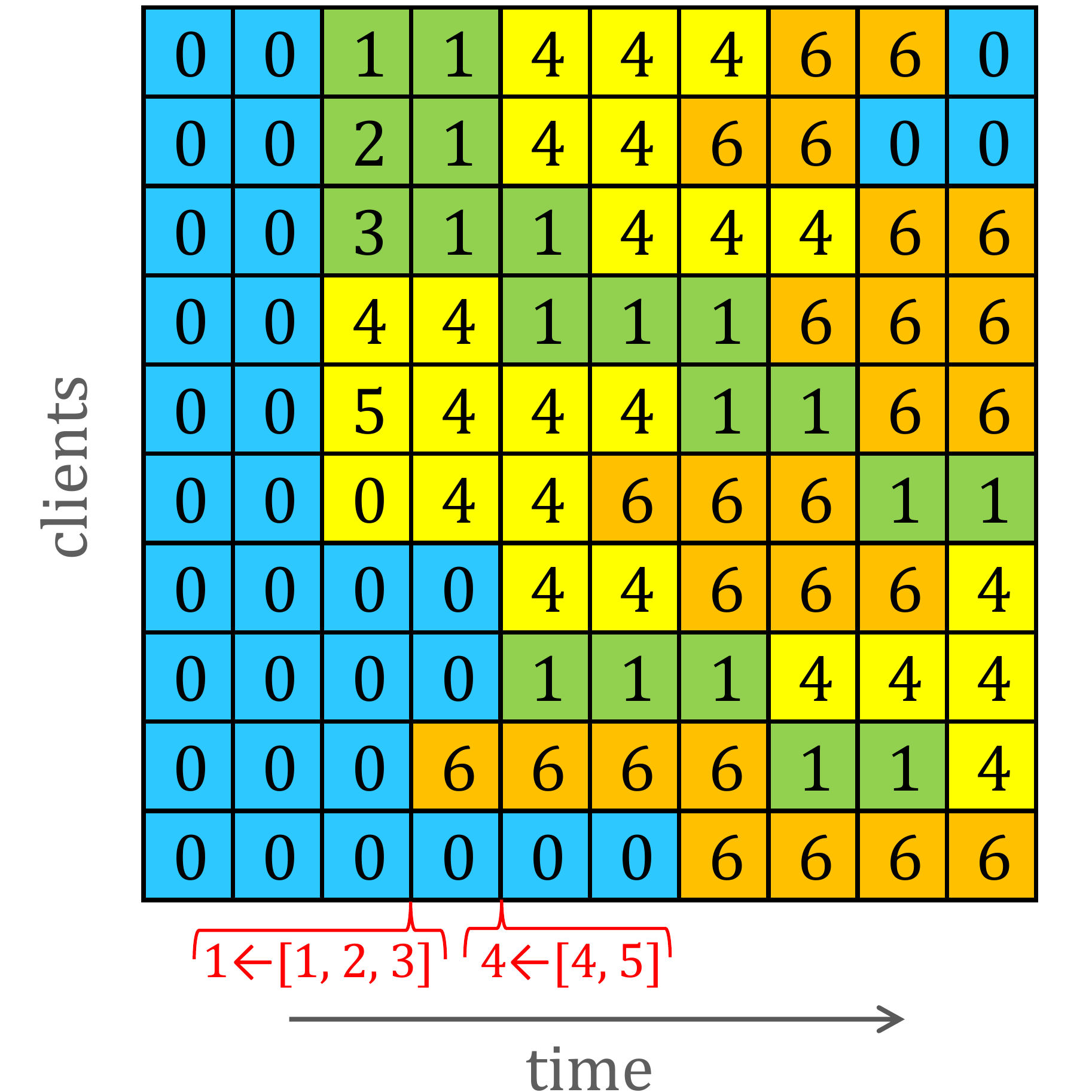}
        \vspace{-0.2in}
\caption{The clustering learned by \name{} on MNIST-4. Each cell indicates the model ID at each client and time step, and the background color indicates the ground-truth concept.}    
\label{fig:mnist-clustering}
\end{wrapfigure}
Regarding the 4-concept drift, Table \ref{table:accuracy} shows that all baselines are ill-suited, while \name performs close to \algfont{Oracle}, and that \mmacc has intermediate performance (due to its false unification of simultaneously emerging concepts).
To understand the performance of \name, see Figure \ref{fig:mnist-clustering}. In the ideal case (\algfont{Oracle}), there would be exactly one model for each concept. For \name, at time 3 one new model is created for 5 of the 6 clients that drifted, and one false negative where a drifted client stays on the original model. With hierarchical clustering applied at the beginning of time 4, the 3 clusters corresponding to the green concept are correctly merged, while all clients on the yellow concept cluster to model 4 which had the lowest test loss over the new data. Also at time 4, model 6 is created for the new orange concept. Then at time 5, hierarchical clustering merges models 4 and 5 (due its iterative application in \name, as the distance decreases after model 4 is further trained). After time 5, \name has a distinct model for each concept, and no excess models.

One drawback of \name is that it can create more models compared to \mmacc, adding to the communication cost. Appendix \ref{sec:appendix-expt-results} shows that restricting \name to just one new global model per time step (additional local models are still permitted) decreases its accuracy by only 0.92\% on the MNIST-4 dataset, while saving communication.

Finally, we discuss the drift in the real-world FMoW dataset where we observe \name has superior performance. The authors of the WILDS benchmark primarily make note of the performance loss of a globally trained model on data from Africa over time \citep{KohSMXZBHYPGLDS21}. We observe \name successfully adapts to the local drift, switching the model applied at Africa at 2014 when there is a significant increase in single-unit residential buildings in Figure \ref{fig:fmow-classes} in \S\ref{sec:motivation}. Instead of creating a new model at 2014, we find \name joins the cluster for Oceania where a local model was previously created, and stays at that cluster for 2014 and 2015, before then splitting into a new individual cluster for 2016 and 2017. We also observe that \name detects a drift at 2015 for both Europe and the Americas, creating two more local models that contribute to higher accuracy. 

Meanwhile, \mmacc similarly adapts to the change in Africa yielding a performance benefit, but it does not adapt well to the simultaneous drift for Europe and the Americas.
Both \name and \mmacc outperform the centralized adaptation baselines which fail to adapt to the drift when viewed globally (c.f. Figure \ref{fig:fmow-classes}). Finally, the low accuracy of \algfont{IFCA} is explained by its random initialization of model parameters for its clusters, in lieu of the pretrained ImageNet initialization under the rest of the algorithms, and the low accuracy of \algfont{KUE} is explained by its ineffective random subspace projections of the data for this task.

\section{DISCUSSION}
\label{sec:discussion}
In this work, we present \mmacc and \name, the first FL solutions designed to address the challenges of distributed concept drifts staggered in time and space (across clients). We empirically confirm the proposed solutions achieve significantly higher accuracy over existing baselines. We discuss the assumptions, limitations and future direction of our work here.

\textbf{Privacy considerations.} The clustering algorithm of {\name} shares the local model learned by a single client with all clients, which could raise privacy concerns. For privacy-sensitive applications, our methods could be combined with other privacy-preserving techniques, e.g., model perturbation \citep[\S4]{kairouz2021advances} in future work.

\textbf{Drift detection methods.} For simplicity, we use a basic drift detection test (Eq (\ref{eq:drift-detect-one}) in \S\ref{sec:algo-creation}) for a change in the loss that exceeds a given threshold. For production use, it would be beneficial to use a state-of-the-art detection test that is more statistically grounded and yields a quantitative statement on the assumption (\S\ref{sec:hierarchical}) that the size of local data samples is large enough for statistical significance when creating and merging clusters. In particular, tests based on loss degradation by a proportional threshold \citep{baena2006early, barros2017rddm} rather than an absolute threshold may be better suited for the multiple-model algorithm (\name), as different models can have different loss magnitudes. We leave the exploration of combining various drift detection tests with our proposed solutions as future work.

\textbf{Concept drifts and anomalies.} We assume all observed concept drifts should be considered. But in the case of anomalies, it may be desirable not to react. One line of related work focuses on adapting only to ``true'' drifts while also exhibiting robustness in the presence of anomalies \citep{togbe2021anomalies, sankararaman2022fitness}. Future work might investigate extending clustering algorithms like \name to include anomaly detection in order to exclude outliers in isolated clusters and prevent false merges that could result in model poisoning.

\textbf{Clustering algorithm alternatives.} A design choice of our clustering algorithms is that we identify each client with the best-performing global model at each time step. An alternative approach is soft-clustering, previously explored by \citet{li2021federated} in the context of static clustering in FL, in which a client fractionally identifies with multiple global models and takes the average for inference. We choose to not use soft-clustering because our preliminary experiments with soft-clustering show no benefit in performance, while increasing communication costs for additional local updates.

\textbf{Model averaging alternatives.} In \name, the initial model parameters for a cluster after merging is the average of the constituent models, weighted by the size of each model's training dataset. An alternative approach to investigate in future work is to use weights that incorporate each model’s loss over a sample of the aggregate dataset (already computed with the $L_{ij}$'s) so that more accurate models are weighted higher, analogous to weighted majority voting.

\section{CONCLUSION}
\label{sec:conclusion}
Federated learning under distributed concept drift is a largely unexplored area, posing particular challenges because drifts can arise staggered in time and space (across clients).
This paper presented \mmacc and \name, the first algorithms explicitly designed to mitigate these challenges.
Empirical evaluation on a variety of dataset/drift combinations showed that these algorithms achieve significantly higher accuracy than existing baselines, and are comparable to an idealized algorithm with oracle knowledge of the ground-truth clustering. %
We hope that our solution spurs further research to this emerging problem, as well as addressing the privacy implications of clustering clients.

\subsubsection*{Acknowledgements}
We thank the anonymous AISTATS reviewers for their valuable and constructive suggestions. This work was supported in part by NSF grants CCF-1919223, CCF-2045694, CNS-2112471, CNS-2211882, ONR N00014-23-1-2149, U.S. Army W911NF20D0002, and a Google Research Collaboration gift award.

\bibliography{drift-fl}

\newpage
\onecolumn
\appendix

\section{DATASETS AND EXPERIMENTAL PARAMETERS}
\label{sec:appendix-expt-setup}

We consider synthetic distributed drifts with respect to the following datasets previously used in the concept drift and personalized FL literature \citep{brzezinski2013reacting, TahmasbiJTG21, briggs2020federated, canonaco2021adaptive, manias2021concept}: SINE and CIRCLE \citep{pesaranghader2016framework} which each have 2 defined concepts, and SEA \citep{bifet2010moa} and MNIST \citep{lecun1998gradient}, which have up to 4 concepts. In SINE, the first concept is a decision boundary of the sine curve $x_2 < \sin(x_1)$ for data points sampled from the unit square, and the second concept reverses the direction (swapping the labels). In CIRCLE, the two concepts are each decision boundaries of two different circles in the unit square, representing a smaller concept change than SINE. The first circle is centered at (0.2, 0.5) with radius 0.15 and the second circle is centered at (0.6, 0.5) with radius 0.25. In SEA, each concept corresponds to a shifted hyperplane. Each point in SEA has three attributes in [0, 10], where the label is determined by $x_1 + x_2 \leq \theta_j$ where $j$ corresponds to 4 concepts, $\theta_A=9, \theta_B=8, \theta_C=7, \theta_D=9.5$. (The third attribute $x_3$ is not correlated with the label.) In SEA, at every concept there is noise in the observed labels, where the label is swapped with 10\% chance for each data point independently. In MNIST, concept A corresponds to the original labeling of the hand-drawn digits, and under each other concept, the labels of two of the digits are swapped (B swaps digits 1 and 2, C swaps digits 3 and 4, and D swaps digits 5 and 6).

For each of the synthetic drift datasets in our experiments, the training data are distributed across 10 clients and arrive over 10 time steps. The partition of the data at each client and time is a constant 500 number of samples from the concept corresponding to the concept drift patterns in Figures \ref{fig:2concepts} and \ref{fig:4concepts} in \S\ref{sec:motivation}. In our experimental results, after training at each time $\tau$ we report the test accuracy over the data at $\tau+1$. For clarification, in reporting the accuracy at the last time step 10, we test over an 11th sample of data at each client that is from the same concept observed during training at time 10.

We also evaluate on the real-world drift in the Functional Map of the World (FMoW) dataset included in the WILDS benchmark \citep{christie2018functional, KohSMXZBHYPGLDS21}. 
The learning task is to classify the land use or building type from satellite images, which has significant practical relevance, ``aiding policy and humanitarian efforts in applications such as deforestation tracking, population density mapping, crop yield prediction, and other economic tracking applications'' \citep{KohSMXZBHYPGLDS21}. Each image is RGB and square with a width of 224 pixels. The WILDS benchmark is not explicitly posed as a drift \emph{adaptation} problem that we study in this paper, but instead as a drift \emph{robustness} problem, and so they originally partitioned the data into train/validation/test splits. For our evaluation, we re-partition the dataset, distributing training data across 5 clients arriving over 9 time steps, using the metadata annotation of each image by region (Africas, Americas, Asia, Europe, Oceania) and year. The first 8 years from 2002--2009 have much fewer images collected, which we group into one time step, and then we treat each year from 2010--2017 as one time step each. The partition of the data at each client and time step is a subsample of up to 1000 images at the 10 classes that are the most common (counting across all regions and years). The test data evaluated for the last time step are a disjoint subsample also from the same year 2017 as the training data. Figure \ref{fig:fmow-classes} in \S\ref{sec:motivation} depicts how the data drifts gradually over time, where the development of new infrastructure is a result of social, political, economic, and environmental factors. Viewed globally, the drift is small. \citet{KohSMXZBHYPGLDS21} write: ``intriguingly, a large subpopulation shift across regions only occurs with a combination of time and region shift.'' Further, they call for solutions that ``can leverage the structure across both space and time'' and also hypothesize a benefit to ``potentially transfer knowledge of other regions with similar economies and infrastructure'' which we empirically confirm where \name clusters Africa and Oceania together for years 2014--2015.

Across all algorithms we evaluate, the algorithms that learn a single model use FedAvg for training, and the clustering algorithms that learn multiple models use Algorithm \ref{alg:training} in \S\ref{sec:algo-training} for training (which reduces to FedAvg when there is one cluster). The training parameters used in our experiments are shown in Table \ref{table:training-params}. For efficiency of the larger FMoW experiments, we reduce to 10 rounds and batch size 32---we observe that this suffices by convergence of the training accuracy. 

\begin{table}[b]
\caption{Training parameters}
\label{table:training-params}
\begin{center}
\begin{tabular}{clll}
  \toprule
  Parameter & Description & Experimental setting  & Experimental setting \\
& & (all synthetic drifts) & (FMoW) \\
  \midrule
    $R$ & \# communication rounds & 100 & 10\\
    $K$ & \# local steps per model per round & 50 & 50\\
    $B$ & minibatch size & 50 & 32\\
    $\eta$ & step size & varies & varies\\
 \bottomrule
  \end{tabular}
  \end{center}
  \end{table}

Regarding the learning rate selection, first we discuss all algorithms excluding \algfont{Adaptive-FedAvg}. We searched for learning rates of the form $10^{-a}$ for $a=1, 2, 3, 4$, for each dataset, and found that $\eta=10^{-2}$ was the best for SINE-2, CIRCLE-2, SEA-2, and SEA-4, that $\eta=10^{-3}$ was best for MNIST-2 and MNIST-4, and that $\eta=10^{-4}$ was best for FMoW. (This held for both of the two extremes among our baselines, \algfont{Oblivious} and \algfont{Oracle}, and we apply the same learning rate across all the algorithms. For FMoW, there is no known \algfont{Oracle}, so we searched only using the \algfont{Oblivious} baseline.) Also note that for computing the \textsc{LocalUpdate} at each client, we use the implementation of Adam in PyTorch with the options weight decay = $10^{-3}$ and amsgrad = True. We treat \algfont{Adaptive-FedAvg} separately, because it uses SGD with its own internal learning rate scheduler as its mechanism to react to drifts. We found that the initial learning rate of $10^{-2}$ was the best for each dataset with the exception of SINE-2, instead using $10^{-1}$. (This higher learning rate explains the high standard deviation in the reported accuracy of \algfont{Adaptive-FedAvg} on SINE-2.)

\begin{figure}[b]
\begin{center}
\vspace{-1.0in}
    \begin{subfigure}[t]{0.25\textwidth}
        \includegraphics[trim=1cm 8cm 0.5cm 7cm, width=\columnwidth]{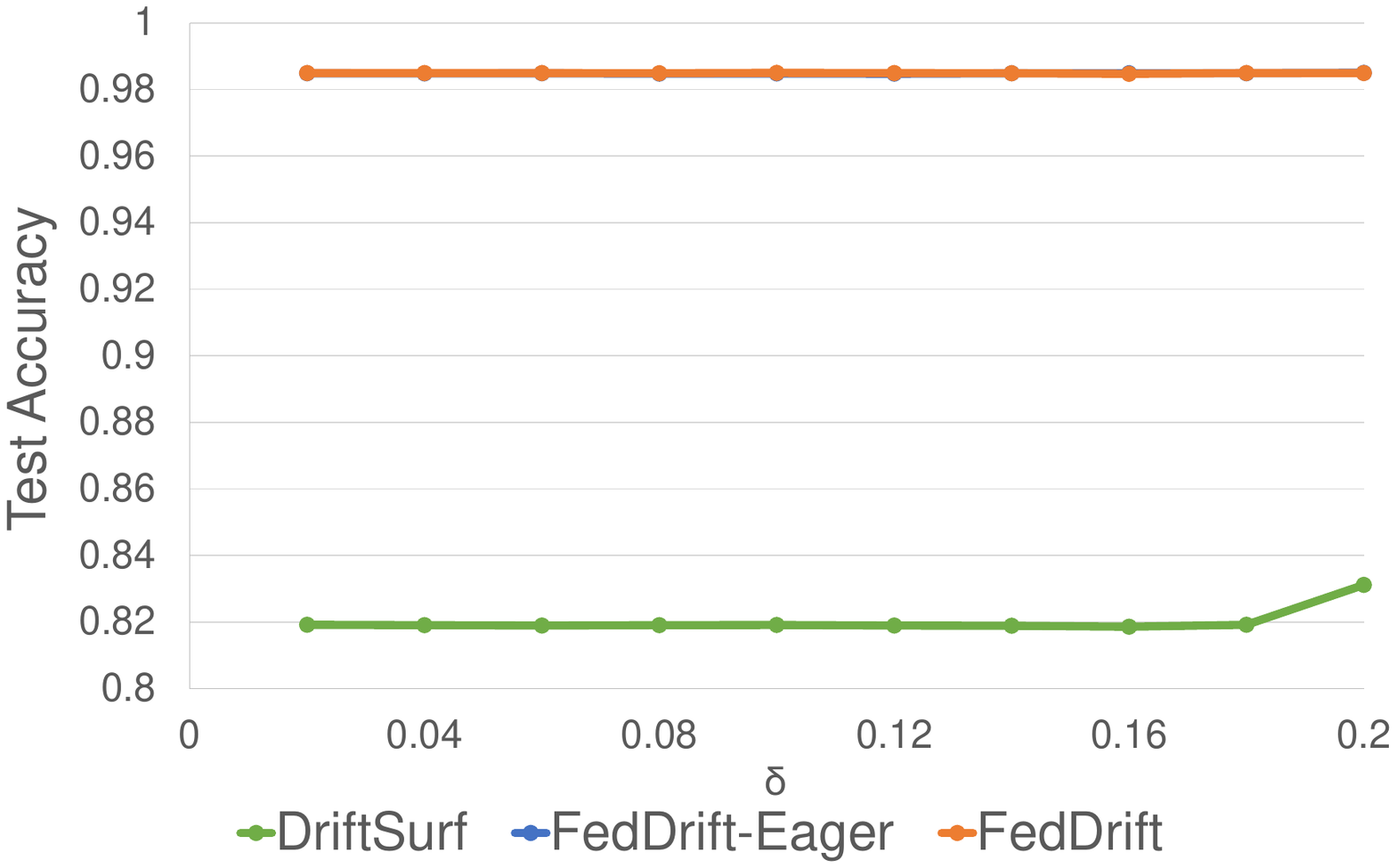}
        \caption{SINE-2}
        \label{fig:delta-sine2}
    \end{subfigure}\hfill
    \begin{subfigure}[t]{0.25\textwidth}
        \includegraphics[trim=1cm 8cm 0.5cm 7cm, width=\columnwidth]{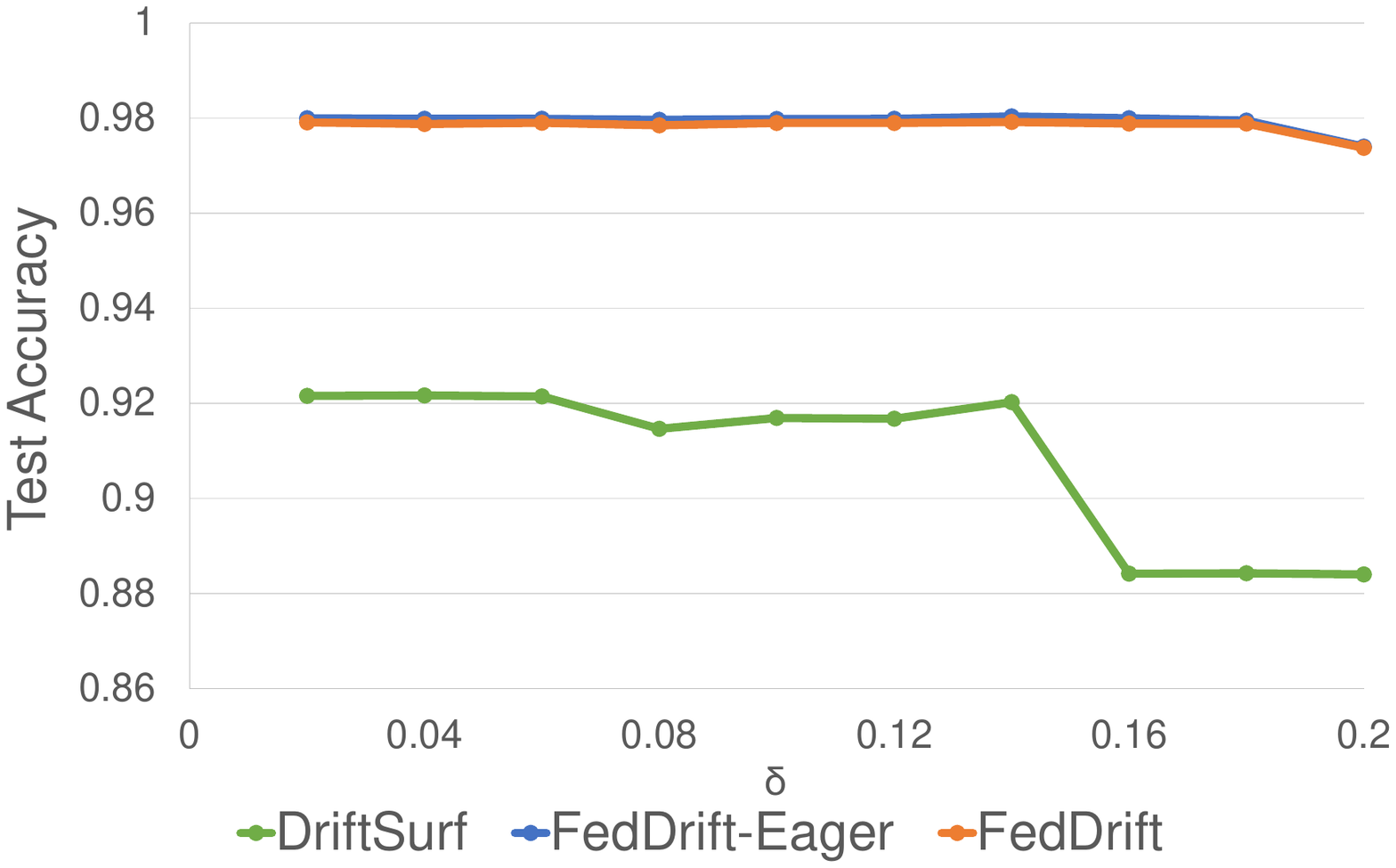}
        \caption{CIRCLE-2}
        \label{fig:delta-circle2}
    \end{subfigure}\hfill 
    \begin{subfigure}[t]{0.25\textwidth}
        \includegraphics[trim=1cm 8cm 0.5cm 7cm, width=\columnwidth]{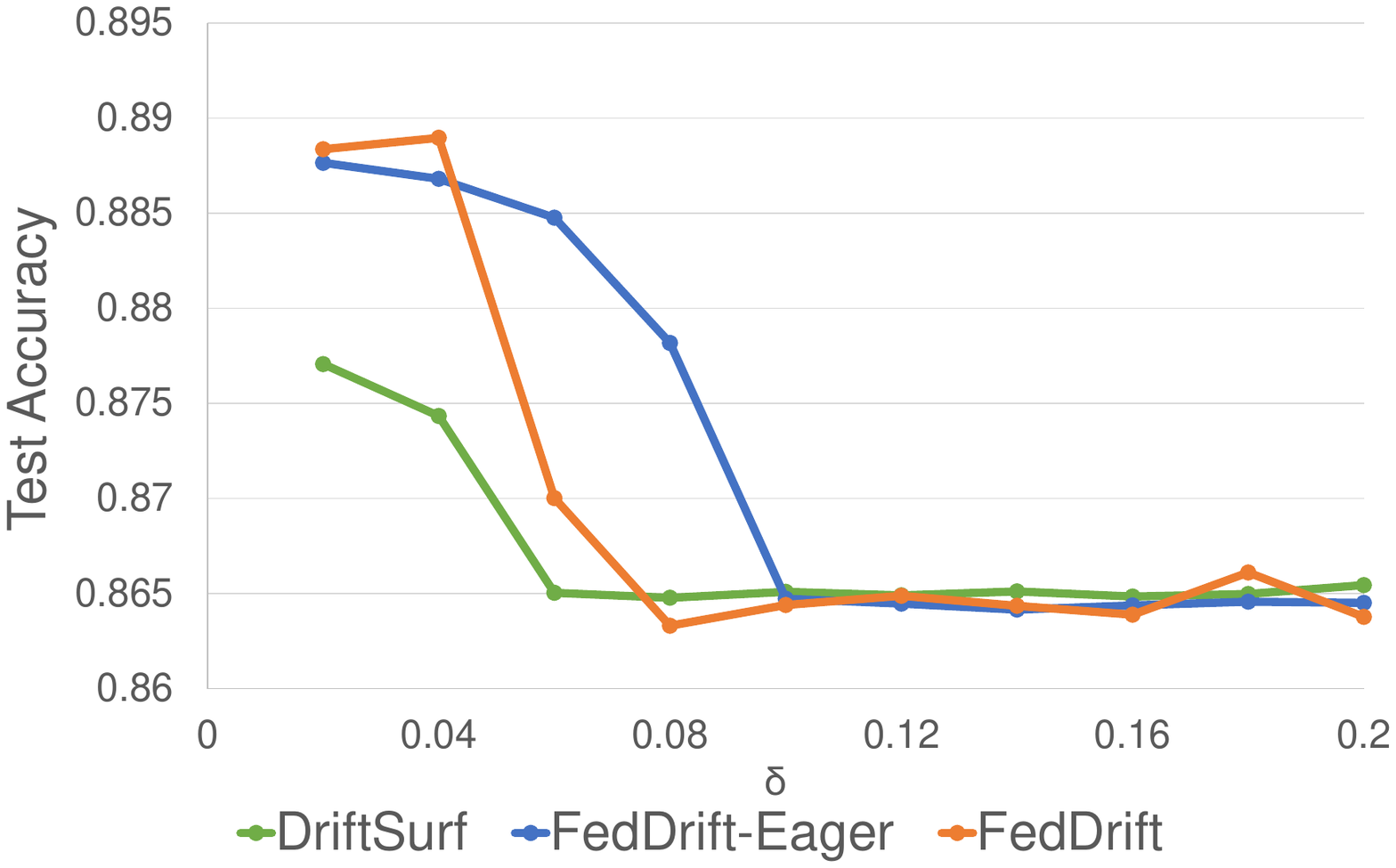}
        \caption{SEA-2}
        \label{fig:delta-sea2}
    \end{subfigure}\hfill
    \begin{subfigure}[t]{0.25\textwidth}
        \includegraphics[trim=1cm 8cm 0.5cm 7cm, width=\columnwidth]{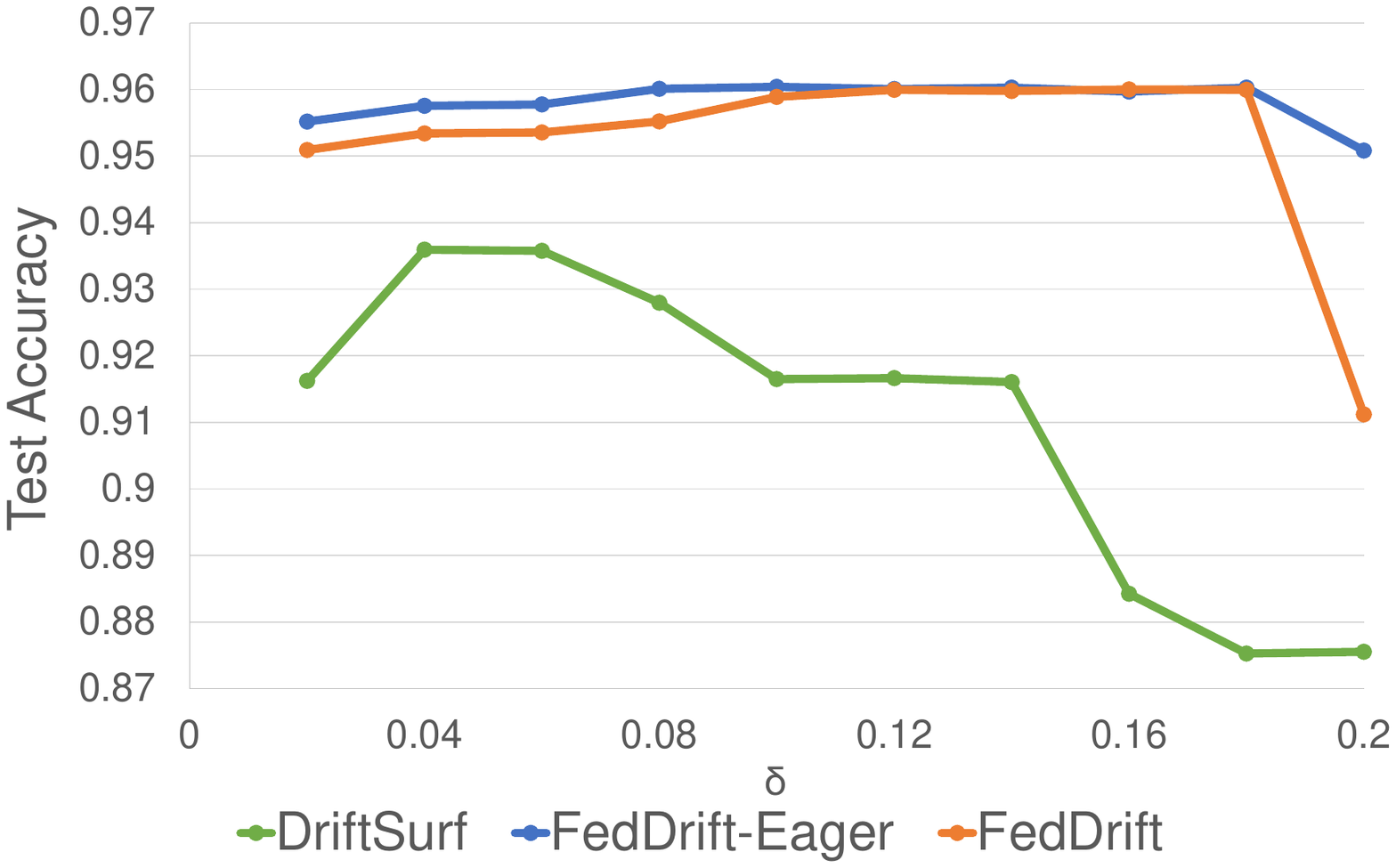}
        \caption{MNIST-2}
        \label{fig:delta-mnist2}
    \end{subfigure}\hfill   
\vspace{+0.1in}    
    \begin{subfigure}[t]{0.25\textwidth}
        \includegraphics[trim=1cm 8cm 0.5cm 7cm, width=\columnwidth]{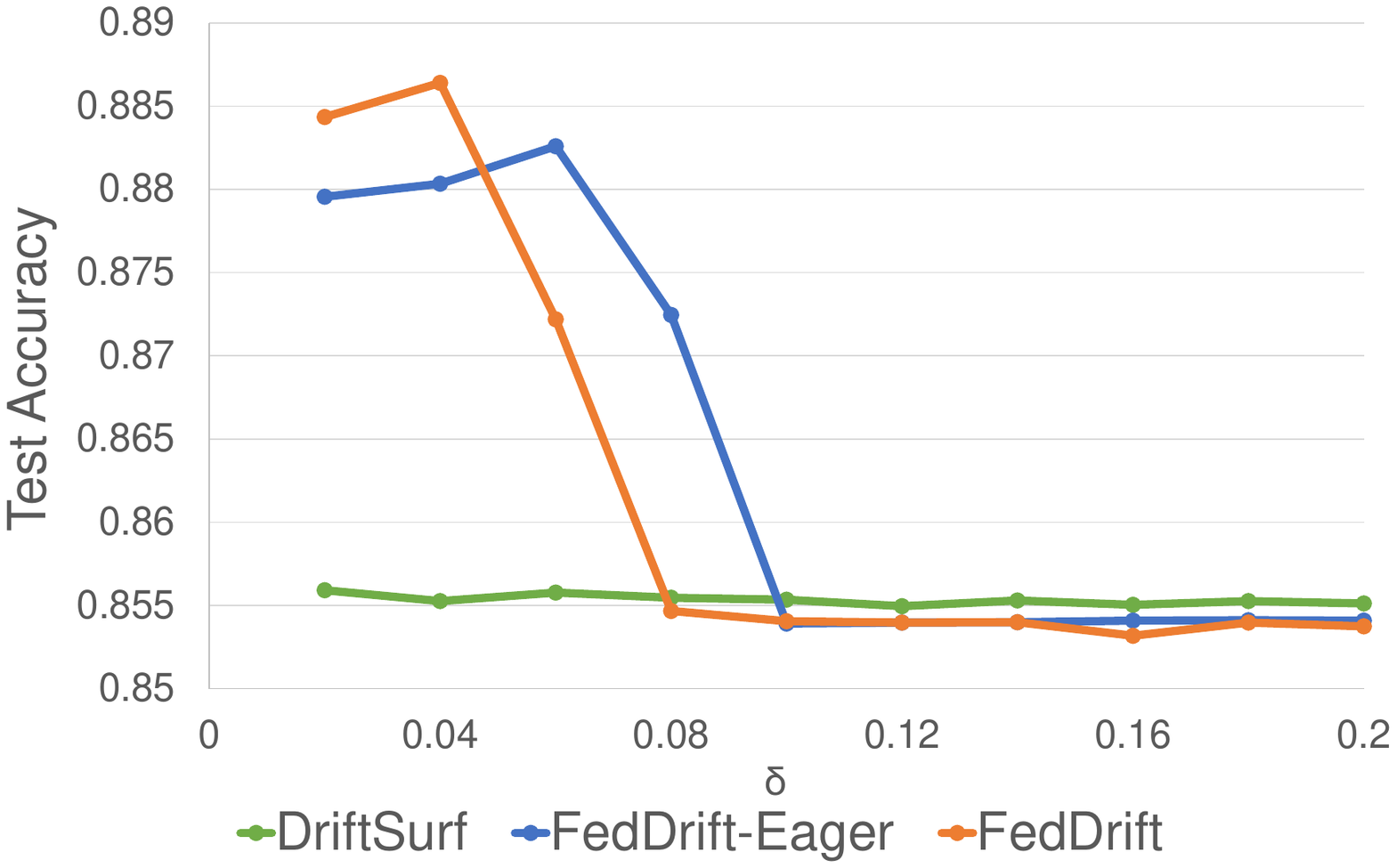}
        \caption{SEA-4}
        \label{fig:delta-sea4}
    \end{subfigure}\hfill
    \begin{subfigure}[t]{0.25\textwidth}
        \includegraphics[trim=1cm 8cm 0.5cm 7cm, width=\columnwidth]{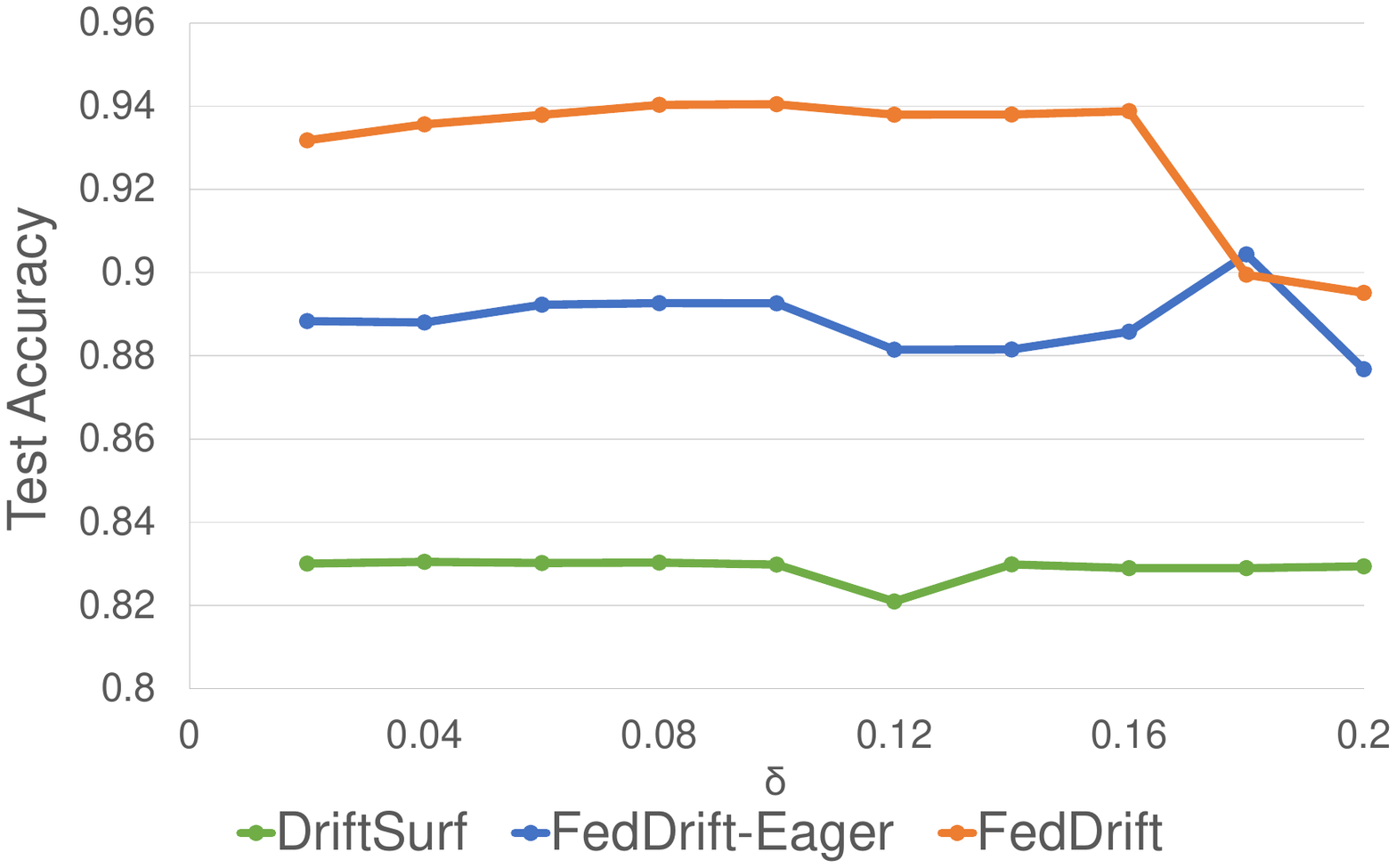}
        \caption{MNIST-4}
        \label{fig:delta-mnist4}
    \end{subfigure}\hfill   
    \begin{subfigure}[t]{0.25\textwidth}
        \includegraphics[trim=1cm 8cm 0.5cm 7cm, width=\columnwidth]{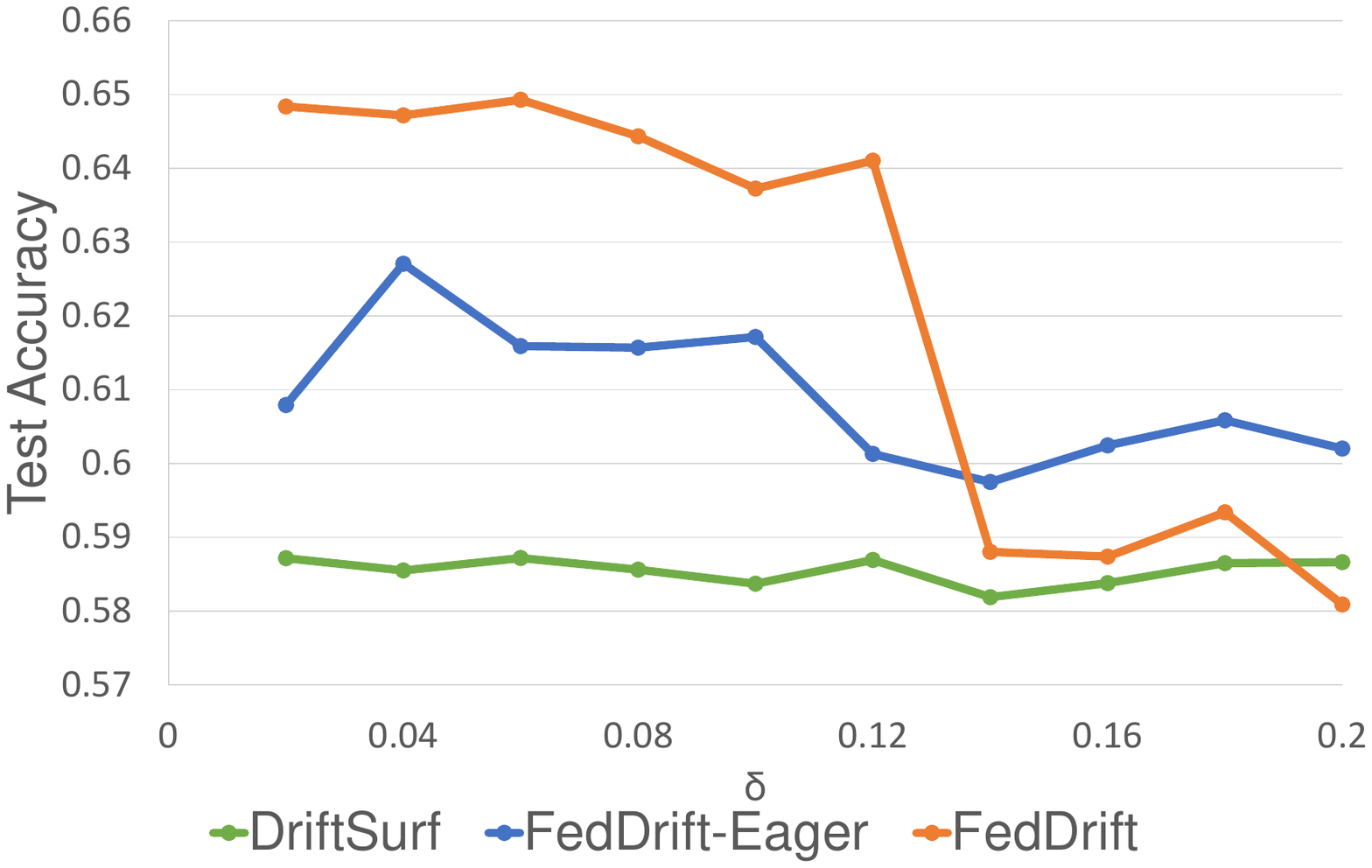}
        \caption{FMoW}
        \label{fig:delta-fmow}
    \end{subfigure}
\end{center}
\caption{Average accuracy of each drift detection-based algorithm under varying thresholds $\delta$.}
\label{fig:delta}
\end{figure}

Next, we report the selection of the drift detection threshold $\delta$ in the algorithms \algfont{DriftSurf}, \mmacc, and \name. While the optimal $\delta$ is expected to vary across datasets, even for a fixed dataset, different algorithms can peak in performance at varying $\delta$. The performance of each of these three algorithms for each dataset across $\delta$ in the range $0.02, 0.04, \dotsc, 0.20$ is shown in Figure \ref{fig:delta}. To not bias towards any one algorithm, the experimental results are reported for each algorithm and dataset using its best $\delta$. (The $\delta$ used for the \algfont{FedDrift-C} variant discussed in Appendix \ref{sec:appendix-expt-results} is identical to that used for \name.) However, using a fixed $\delta=0.04$ for \mmacc and \name makes at most a 1 pp difference in the results reported in Table \ref{table:accuracy} (on one trial).

For all other hyperparameters of the algorithms we evaluate, we follow the parameter choices stated in the original papers, with the following exceptions: for \algfont{DriftSurf} we use $r=3$ (which performed better than their suggested $r=4$); for \algfont{CFL} we use $\gamma = 0.1$ (for which there is no default, but is shown to be a good setting from Theorem 1 and Figure 3 of their paper \cite{sattler2020clustered} given that the number of distinct concepts at a time is at most 5 across all evaluated datasets); and for \algfont{AUE} we use $K=5$ as the total ensemble size (compared to the $K=10$ in their paper they consider over a significantly longer time horizon). In reporting FMoW results, for training efficiency, we further restrict to a total ensemble size of 4 for \algfont{AUE} and \algfont{KUE}.

Furthermore, for the FMoW dataset, which has more than one distinct data distribution at the initial time step unlike the remaining datasets, we use a different initialization of \algfont{IFCA} variants and \name. For \algfont{IFCA} variants, clients initially self-select among 5 cluster centers instead of being all assigned to a single cluster. For \name, clients are initialized to a local model each, which can be merged starting at the next time step. (If we instead initialize all clients to a single model that can later be split, we observed the average test accuracy of \name is 64.46\%, or 0.38\% worse.)

Finally, regarding the model training in Algorithm \ref{alg:training} at time $\tau$, we apply one optimization for efficiency to only train models that are currently clustered to. (Although note that any such models are still retained by \mmacc and \name in order to react to recurring drifts even if they are not actively being trained.)

\FloatBarrier
\section{ADDITIONAL EXPERIMENTAL RESULTS}
\label{sec:appendix-expt-results}

We present additional experimental results on more baseline algorithms and on variants of our algorithms restricted to limited memory or communication.

\paragraph{Additional Baseline Algorithms.}
The additional algorithms presented in this appendix are:
\begin{itemize}
\item \textbf{Four traditional drift adaptation algorithms.} \algfont{AUE-PC} is a variation of the ensemble method \algfont{AUE} with the ensemble weights set \emph{per-client}. \algfont{Window-2} is a window method like \algfont{Window}, except that it forgets data older than two time steps instead of one. \algfont{Weighted-Linear} and \algfont{Weighted-Exp} also forget older data like window methods, but do so more gradually by down-weighting older data with either linear or exponential decay.
\item \textbf{The FL clustering algorithm \algfont{CFL} \citep{sattler2020clustered}.} In extending the original static algorithm to our time-varying setting, we also consider a variant \algfont{CFL-W}, in which during training, each client samples only from the window of the newest data arriving at each time.
\item \textbf{Three variations of the \algfont{IFCA} clustering algorithm \citep{ghosh2020efficient}} that we considered for extending the original algorithm to the time-varying setting. First, \algfont{IFCA(T)} is exactly Algorithm \ref{alg:clustering-loss} in \S\ref{sec:algo-training}, which defines cluster identities for each client and each time, in order to associate the data within a client that are heterogeneous over time across multiple clusters. \algfont{IFCA(T)} chooses the cluster identity once per \emph{time step} (where time steps consist of multiple communication rounds)---this differs from the original algorithm described by \citet{ghosh2020efficient}, which recomputes the cluster identity once per \emph{round}. Second, \algfont{IFCA} does the per-round clustering; more precisely, for each time step $\tau$, the cluster identity $w_{c,m}^{(\tau)}$ is recomputed at every round under the same equation used at the beginning of the time step in Algorithm \ref{alg:clustering-loss}. Third, \algfont{IFCA-W} is a variant of \algfont{IFCA} that trains only over the most recent data arrivals at each time, and the cluster identities of data from previous time steps are forgotten. In general, the \algfont{IFCA}-based algorithms require the number of clusters as input, which we  provide as oracle knowledge---either 2 or 4 depending on the total number of concepts over time in each dataset. This gives \algfont{IFCA}-based algorithms an advantage over all other algorithms we evaluate, which do not know the number of clusters a priori. For the initialization of all three variations, at time 1 and round 1, all clients are assigned to a single cluster, matching the assumption we made for \name and \mmacc in \S\ref{sec:algo-creation}. The exception to this initialization strategy is on FMoW, where the total number of concepts is not known, and the concept at time 1 across clients is not identical; for this dataset, we instead initialize all \algfont{IFCA}-based algorithms with a total of 5 clusters (matching the number of regions), and where each client identifies with the best-performing randomly initialized model (same as the original paper).
\item \textbf{A more communication-efficient variant of \name.} \algfont{FedDrift-C} is the algorithm referred to in the last paragraph of \S\ref{sec:algo-creation} that is restricted to introducing one new global model per time step. More details on this algorithm are described later in this section.
\item \textbf{Sliding window variants of \mmacc and \name.} \algfont{FedDrift-Eager-W} and \algfont{FedDrift-W} are restricted to using only the most recent time step of data $S_c^{(t)}$ and cluster identities $w_{c, m}^{(t)}$.
\item \textbf{A baseline sliding window variant \algfont{Oracle-W}}, which has oracle access to the ground-truth clustering but only uses the most recent time step of data in training.
\end{itemize}

In general, we use the \algfont{-W} suffix in the name of an algorithm to indicate a limited memory of a window of one time step. This memory restriction reduces the number of samples used for training at a time and might reduce the accuracy achievable under ground-truth clustering (\algfont{Oracle-W} vs. \algfont{Oracle}). Yet, the window is not strictly a drawback: (i) forgetting the older data builds in a passive adaptation to drift and (ii) in our setting it also guarantees that each client's training data at a step are all drawn from the same distribution---this is why we also investigate \algfont{-W} variants when extending the prior static clustering algorithms \algfont{CFL} and \algfont{IFCA} to our setting when data arrive over time.

\paragraph{Test Accuracy Results.}
Table \ref{table:accuracy-extra} (extending Table \ref{table:accuracy} in \S\ref{sec:expt}) shows the test accuracy of all algorithms, averaged across all clients and time steps, but omitting the times of drifts. As noted in \S\ref{sec:expt}, we omit the times of drift when all algorithms suffer from the performance loss. For completeness, the test accuracy averaged over all time steps including drifts is shown in Table \ref{table:accuracy-drift-extra}. 
In this latter table, note that \algfont{Oracle} and \algfont{Oracle-W} suffer a performance loss too at the time of drift. Under the test-then-train evaluation, \algfont{Oracle} has access to the concept ID of the data at training time but not at test time, where at each client, the model used for inference corresponds to the observed concept in the most recently arrived training data.
Note that for the real-world gradual drifts in FMoW, the ground-truth is unknown, so we omit results for \algfont{Oracle}. Furthermore, because drifts occur gradually and there is no oracle knowledge of their timing, we report identical test accuracy results on FMoW in Tables \ref{table:accuracy-extra} and \ref{table:accuracy-drift-extra}, averaging across all clients and time steps.

\begin{table}[t]
\setlength{\tabcolsep}{4pt}
\caption{Average test accuracy (\%) across clients and time, omitting drifts (5 trials)}
\label{table:accuracy-extra}
\begin{center}
\begin{small}
\begin{tabular}{lccccccc}
\toprule
 & SINE-2 & CIRCLE-2 & SEA-2 & MNIST-2 & SEA-4 & MNIST-4 & FMoW \\
\midrule 	
\algfont{Oblivious}	&	52.11	$\pm$	1.79	&	88.38	$\pm$	0.17	&	86.46	$\pm$	0.22	&	87.37	$\pm$	0.16	&	85.40	$\pm$	0.09	&	82.95	$\pm$	0.03	&	58.57	$\pm$	0.07	\\
\algfont{DriftSurf}	&	84.18	$\pm$	1.40	&	92.34	$\pm$	0.38	&	87.20	$\pm$	0.27	&	93.26	$\pm$	0.52	&	85.55	$\pm$	0.13	&	82.97	$\pm$	0.09	&	58.45	$\pm$	0.19	\\
\algfont{KUE}	&	86.86	$\pm$	0.17	&	93.71	$\pm$	0.14	&	87.25	$\pm$	0.94	&	90.44	$\pm$	0.44	&	85.09	$\pm$	0.86	&	79.89	$\pm$	0.26	&	33.11	$\pm$	6.09	\\
\algfont{AUE}	&	86.00	$\pm$	0.95	&	92.84	$\pm$	0.19	&	87.48	$\pm$	0.07	&	92.22	$\pm$	0.05	&	85.47	$\pm$	0.12	&	82.07	$\pm$	0.47	&	54.23	$\pm$	0.14	\\
\algfont{AUE-PC}	&	88.67	$\pm$	0.73	&	92.82	$\pm$	0.23	&	87.55	$\pm$	0.20	&	92.24	$\pm$	0.03	&	86.67	$\pm$	0.05	&	81.84	$\pm$	0.33	&	54.15	$\pm$	0.10	\\
\algfont{Window}	&	86.28	$\pm$	0.64	&	93.72	$\pm$	0.14	&	87.94	$\pm$	0.10	&	92.34	$\pm$	0.07	&	85.72	$\pm$	0.13	&	81.43	$\pm$	0.44	&	58.88	$\pm$	0.15	\\
\algfont{Window-2}	&	85.97	$\pm$	0.94	&	93.28	$\pm$	0.15	&	87.62	$\pm$	0.33	&	92.80	$\pm$	0.44	&	86.58	$\pm$	0.15	&	82.22	$\pm$	0.30	&	59.44	$\pm$	0.23	\\
\algfont{Weighted-Linear}	&	72.93	$\pm$	2.05	&	89.87	$\pm$	0.54	&	87.02	$\pm$	0.46	&	89.86	$\pm$	0.36	&	86.42	$\pm$	0.10	&	82.74	$\pm$	0.04	&	58.05	$\pm$	0.17	\\
\algfont{Weighted-Exp}	&	82.52	$\pm$	1.78	&	92.38	$\pm$	0.32	&	87.11	$\pm$	0.34	&	92.52	$\pm$	0.20	&	86.60	$\pm$	0.07	&	82.51	$\pm$	0.09	&	58.49	$\pm$	0.09	\\
\algfont{Adaptive-FedAvg}	&	74.10	$\pm$	10.03	&	86.26	$\pm$	0.00	&	86.77	$\pm$	0.53	&	92.18	$\pm$	0.05	&	85.25	$\pm$	0.27	&	81.64	$\pm$	0.04	&	52.82	$\pm$	0.21	\\
\algfont{CFL} 	&	57.57	$\pm$	8.87	&	86.59	$\pm$	3.42	&	86.46	$\pm$	0.24	&	86.54	$\pm$	0.43	&	86.24	$\pm$	0.15	&	80.97	$\pm$	0.78	&	57.92	$\pm$	0.32	\\
\algfont{CFL-W}	&	96.92	$\pm$	1.84	&	96.04	$\pm$	1.56	&	87.81	$\pm$	0.32	&	90.66	$\pm$	0.35	&	86.06	$\pm$	0.11	&	80.51	$\pm$	0.72	&	58.82	$\pm$	0.11	\\
\algfont{IFCA(T)}	&	98.45	$\pm$	0.03	&	91.72	$\pm$	5.19	&	86.46	$\pm$	0.23	&	87.33	$\pm$	0.15	&	85.41	$\pm$	0.20	&	82.90	$\pm$	0.05	&	47.76	$\pm$	1.98	\\
\algfont{IFCA}	&	98.46	$\pm$	0.02	&	92.20	$\pm$	5.32	&	86.45	$\pm$	0.25	&	87.55	$\pm$	0.25	&	85.35	$\pm$	0.09	&	82.89	$\pm$	0.04	&	48.17	$\pm$	1.30	\\
\algfont{IFCA-W}	&	\textbf{98.49	$\pm$	0.13}	&	94.31	$\pm$	1.62	&	\textbf{88.04	$\pm$	0.17}	&	91.76	$\pm$	0.50	&	86.17	$\pm$	1.00	&	81.27	$\pm$	0.43	&	49.40	$\pm$	0.76	\\
\midrule																													
\mmacc	&	97.53	$\pm$	0.13	&	\textbf{97.82	$\pm$	0.17}	&	87.51	$\pm$	0.88	&	\textbf{95.52	$\pm$	0.11}	&	87.61	$\pm$	1.26	&	90.69	$\pm$	1.20	&	61.77	$\pm$	0.51	\\
\name 	&	97.43	$\pm$	0.06	&	\textbf{97.82	$\pm$	0.19}	&	87.29	$\pm$	0.75	&	95.48	$\pm$	0.08	&	88.13	$\pm$	0.76	&	\textbf{93.80	$\pm$	0.08}	&	\textbf{64.84	$\pm$	0.33}	\\
\algfont{FedDrift-C}	&	97.91	$\pm$	0.70	&	97.61	$\pm$	0.19	&	87.52	$\pm$	0.91	&	95.45	$\pm$	0.13	&	88.26	$\pm$	0.80	&	92.88	$\pm$	0.39	&	61.86	$\pm$	0.30	\\
\algfont{FedDrift-Eager-W}	&	97.95	$\pm$	0.67	&	97.56	$\pm$	0.24	&	87.32	$\pm$	1.02	&	93.41	$\pm$	1.14	&	86.99	$\pm$	0.40	&	89.59	$\pm$	0.38	&	61.94	$\pm$	0.38	\\
\algfont{FedDrift-W}	&	97.86	$\pm$	0.59	&	97.52	$\pm$	0.22	&	87.30	$\pm$	1.01	&	93.85	$\pm$	0.06	&	\textbf{88.56	$\pm$	0.39}	&	91.34	$\pm$	0.06	&	64.22	$\pm$	0.60	\\
\midrule																													
\algfont{Oracle}	&	98.45	$\pm$	0.03	&	97.84	$\pm$	0.22	&	87.76	$\pm$	0.98	&	95.54	$\pm$	0.11	&	88.79	$\pm$	0.41	&	94.30	$\pm$	0.08	&	-			\\
\algfont{Oracle-W}	&	98.53	$\pm$	0.15	&	97.81	$\pm$	0.13	&	87.31	$\pm$	0.75	&	93.91	$\pm$	0.05	&	88.41	$\pm$	0.57	&	91.75	$\pm$	0.05	&	-			\\
\bottomrule
\end{tabular}
\end{small}
\end{center}
\end{table}

\begin{table}[t]
\setlength{\tabcolsep}{4pt}
\caption{Average test accuracy (\%) across clients and time, including drifts (5 trials)}
\label{table:accuracy-drift-extra}
\begin{center}
\begin{small}
\begin{tabular}{lccccccc}
\toprule
 & SINE-2 & CIRCLE-2 & SEA-2 & MNIST-2 & SEA-4 & MNIST-4 & FMoW \\
\midrule
\algfont{Oblivious}	&	47.36	$\pm$	1.74	&	87.15	$\pm$	0.15	&	86.22	$\pm$	0.21	&	86.40	$\pm$	0.15	&	85.16	$\pm$	0.06	&	81.59	$\pm$	0.02	&	58.57	$\pm$	0.07	\\
\algfont{DriftSurf}	&	79.45	$\pm$	1.55	&	90.98	$\pm$	0.36	&	86.91	$\pm$	0.27	&	92.24	$\pm$	0.64	&	85.19	$\pm$	0.16	&	81.59	$\pm$	0.05	&	58.45	$\pm$	0.19	\\
\algfont{KUE}	&	82.56	$\pm$	0.18	&	92.45	$\pm$	0.12	&	87.02	$\pm$	0.92	&	89.59	$\pm$	0.58	&	84.81	$\pm$	0.68	&	77.84	$\pm$	0.30	&	33.11	$\pm$	6.09	\\
\algfont{AUE}	&	81.24	$\pm$	1.29	&	91.60	$\pm$	0.17	&	87.23	$\pm$	0.07	&	91.09	$\pm$	0.04	&	85.09	$\pm$	0.07	&	79.95	$\pm$	0.63	&	54.23	$\pm$	0.14	\\
\algfont{AUE-PC}	&	83.65	$\pm$	0.92	&	91.58	$\pm$	0.21	&	87.38	$\pm$	0.18	&	91.15	$\pm$	0.06	&	86.30	$\pm$	0.10	&	79.58	$\pm$	0.47	&	54.15	$\pm$	0.10	\\
\algfont{Window}	&	81.77	$\pm$	0.66	&	92.46	$\pm$	0.12	&	87.72	$\pm$	0.09	&	91.58	$\pm$	0.07	&	85.30	$\pm$	0.09	&	78.84	$\pm$	0.26	&	58.88	$\pm$	0.15	\\
\algfont{Window-2}	&	81.46	$\pm$	0.93	&	92.00	$\pm$	0.15	&	87.43	$\pm$	0.38	&	91.79	$\pm$	0.56	&	86.18	$\pm$	0.16	&	79.96	$\pm$	0.49	&	59.44	$\pm$	0.23	\\
\algfont{Weighted-Linear}	&	67.34	$\pm$	1.92	&	88.59	$\pm$	0.52	&	86.77	$\pm$	0.51	&	88.74	$\pm$	0.36	&	86.13	$\pm$	0.13	&	81.31	$\pm$	0.04	&	58.05	$\pm$	0.17	\\
\algfont{Weighted-Exp}	&	76.86	$\pm$	1.82	&	91.03	$\pm$	0.31	&	86.91	$\pm$	0.34	&	91.38	$\pm$	0.20	&	86.26	$\pm$	0.11	&	80.91	$\pm$	0.09	&	58.49	$\pm$	0.09	\\
\algfont{Adaptive-FedAvg}	&	69.69	$\pm$	10.13	&	85.60	$\pm$	0.00	&	86.62	$\pm$	0.50	&	91.33	$\pm$	0.05	&	84.95	$\pm$	0.26	&	79.49	$\pm$	0.04	&	52.82	$\pm$	0.21	\\
\algfont{CFL} 	&	51.98	$\pm$	8.01	&	85.33	$\pm$	3.35	&	86.19	$\pm$	0.29	&	85.54	$\pm$	0.41	&	85.95	$\pm$	0.22	&	79.36	$\pm$	0.94	&	57.92	$\pm$	0.32	\\
\algfont{CFL-W}	&	87.65	$\pm$	1.27	&	94.00	$\pm$	1.32	&	87.56	$\pm$	0.32	&	89.73	$\pm$	0.30	&	85.38	$\pm$	0.14	&	78.15	$\pm$	1.13	&	58.82	$\pm$	0.11	\\
\algfont{IFCA(T)}	&	88.77	$\pm$	0.02	&	90.06	$\pm$	4.62	&	86.22	$\pm$	0.22	&	86.36	$\pm$	0.14	&	85.10	$\pm$	0.13	&	81.53	$\pm$	0.05	&	47.76	$\pm$	1.98	\\
\algfont{IFCA}	&	88.78	$\pm$	0.02	&	90.49	$\pm$	4.73	&	86.21	$\pm$	0.28	&	86.56	$\pm$	0.21	&	85.06	$\pm$	0.04	&	81.51	$\pm$	0.03	&	48.17	$\pm$	1.30	\\
\algfont{IFCA-W}	&	\textbf{88.80	$\pm$	0.12}	&	92.84	$\pm$	1.19	&	\textbf{87.84	$\pm$	0.14}	&	90.81	$\pm$	0.67	&	85.52	$\pm$	0.50	&	79.17	$\pm$	0.39	&	49.40	$\pm$	0.76	\\
\midrule																													
\mmacc	&	87.93	$\pm$	0.12	&	\textbf{95.50	$\pm$	0.14}	&	87.01	$\pm$	0.72	&	\textbf{93.63	$\pm$	0.10}	&	86.73	$\pm$	0.64	&	83.99	$\pm$	0.72	&	61.77	$\pm$	0.51	\\
\name 	&	87.84	$\pm$	0.05	&	\textbf{95.50	$\pm$	0.15}	&	86.85	$\pm$	0.60	&	93.60	$\pm$	0.07	&	86.95	$\pm$	0.51	&	\textbf{85.44	$\pm$	0.08}	&	\textbf{64.84	$\pm$	0.33}	\\
\algfont{FedDrift-C}	&	88.27	$\pm$	0.61	&	95.30	$\pm$	0.17	&	87.04	$\pm$	0.70	&	93.58	$\pm$	0.13	&	\textbf{86.98	$\pm$	0.52}	&	85.30	$\pm$	0.43	&	61.86	$\pm$	0.30	\\
\algfont{FedDrift-Eager-W}	&	88.31	$\pm$	0.59	&	95.23	$\pm$	0.23	&	86.90	$\pm$	0.91	&	91.84	$\pm$	0.16	&	86.50	$\pm$	0.25	&	81.97	$\pm$	0.21	&	61.94	$\pm$	0.38	\\
\algfont{FedDrift-W}	&	88.22	$\pm$	0.52	&	95.20	$\pm$	0.21	&	86.95	$\pm$	0.89	&	91.85	$\pm$	0.06	&	\textbf{86.98	$\pm$	0.36}	&	83.14	$\pm$	0.06	&	64.22	$\pm$	0.60	\\
\midrule																													
\algfont{Oracle}	&	88.76	$\pm$	0.02	&	95.51	$\pm$	0.18	&	87.23	$\pm$	0.93	&	93.65	$\pm$	0.10	&	86.99	$\pm$	0.40	&	85.81	$\pm$	0.07	&	-			\\
\algfont{Oracle-W}	&	88.83	$\pm$	0.14	&	95.48	$\pm$	0.11	&	86.89	$\pm$	0.63	&	91.91	$\pm$	0.05	&	86.56	$\pm$	0.70	&	83.46	$\pm$	0.03	&	-			\\
\bottomrule
\end{tabular}
\end{small}
\end{center}
\end{table}

Based on these tables, we make the following observations on the additional algorithms. The \algfont{AUE-PC} variant of \algfont{AUE} extends the model weights in the ensemble method to be individualized per-client, based on the performance of each model over each client's local data (as opposed to weights chosen based on the aggregate performance at the server). This additional flexibility leads to only a marginal accuracy improvement over \algfont{AUE} across all datasets. While it is generally valuable for clients at different stages of a staggered drift to use different models for inference, the more fundamental obstacle is that each global model trained by \algfont{AUE-PC} is updated by all clients. In the course of the 2-concept staggered drift, all of the models in the ensemble are trained either over a mixture of data from both concepts or solely from the first concept, and there is no accurate model available that is a good fit for the second concept.

The \algfont{Window-2} algorithm and the weighted sampling algorithms \algfont{Weighted-Linear} and \algfont{Weighted-Exp} are techniques for forgetting older data, but less abruptly compared to \algfont{Window-1}, and in general they all perform similarly. On the sharp drift of SINE-2, the fastest forgetting algorithm \algfont{Window} performs the best of these. On the other hand, on the 4-concept drift of MNIST-4 in which the time axis does not well separate different concepts, the slowest forgetting algorithm \algfont{Weighted-Linear} performs best. Meanwhile, the performance of all four algorithms are close on the SEA datasets, which have greater overlap between the concepts.

\begin{figure}[ht!]
\begin{minipage}{0.45\linewidth}
\centering
        \includegraphics[width=0.45\linewidth]{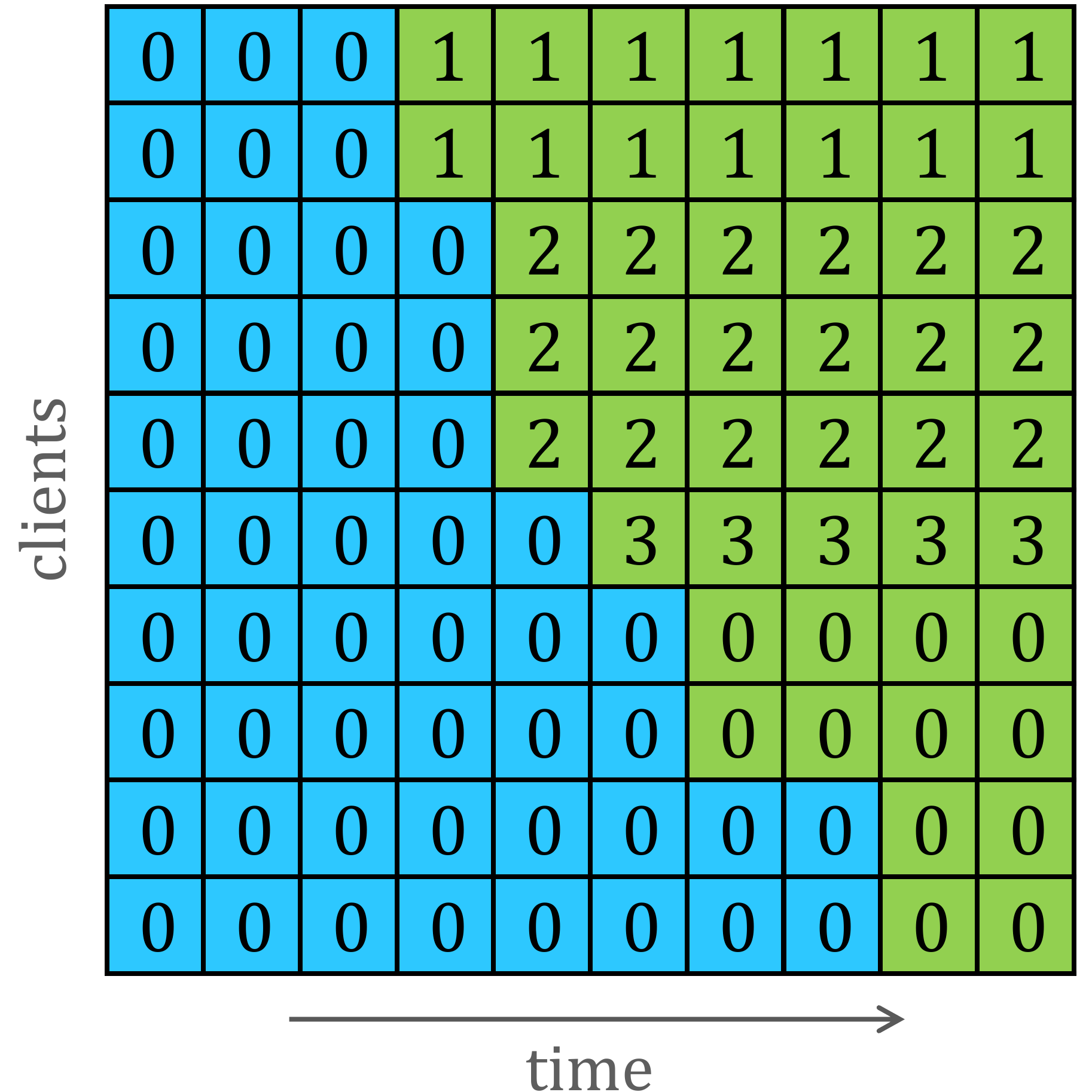}
        \caption{The clustering learned by \algfont{CFL-W} on SINE-2. Each cell indicates the model ID at each client and time step, and the background color indicates the ground-truth concept.}
        \label{fig:cfl}
\end{minipage}
\hfill
\begin{minipage}{0.45\linewidth}
\centering
        \includegraphics[width=0.45\linewidth]{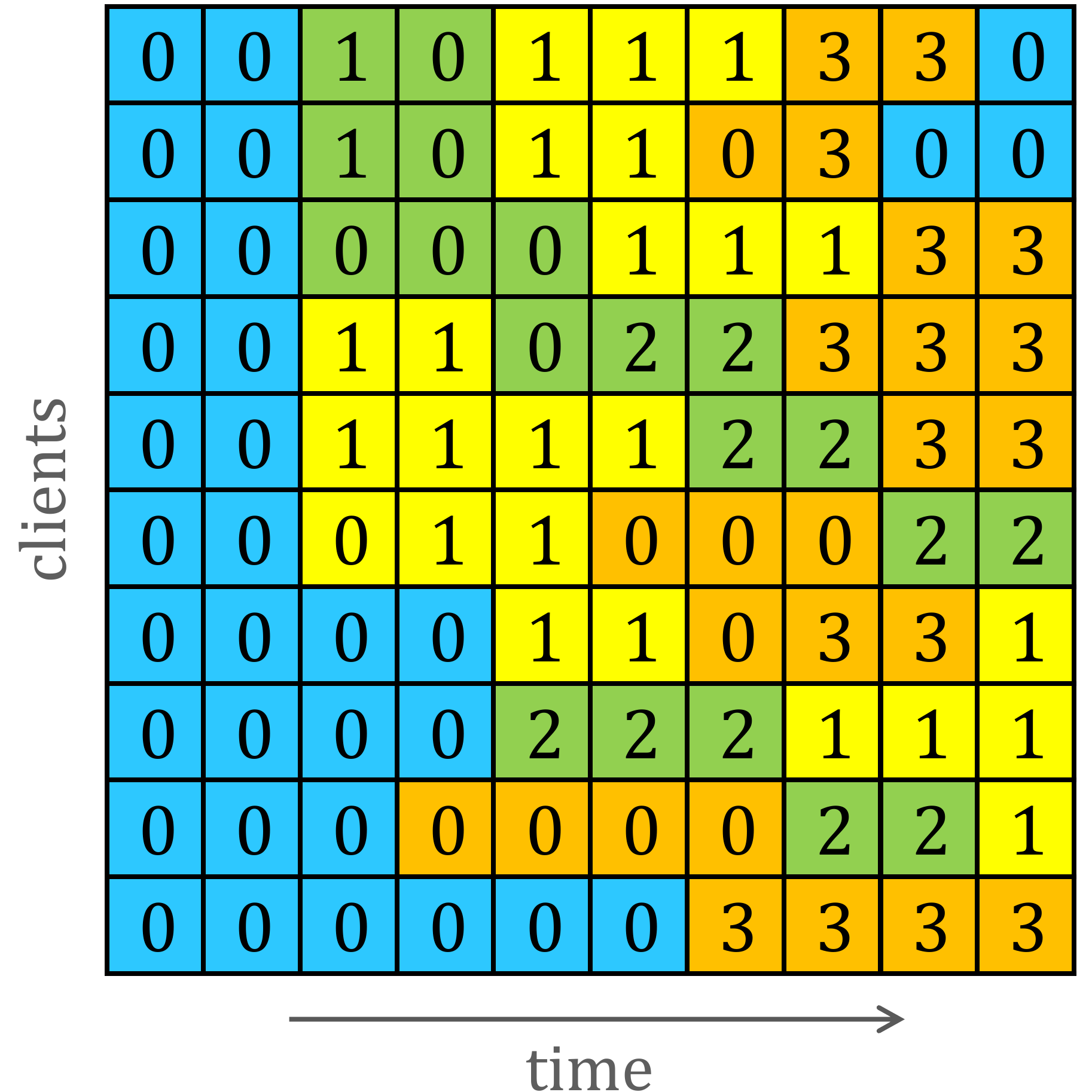}
        \caption{The clustering learned by \mmacc on MNIST-4. Each cell indicates the model ID at each client and time step, and the background color indicates the ground-truth concept.}
        \label{fig:mmacc-mnist4}
\end{minipage}
\end{figure}

The clustering algorithms \algfont{CFL} and \algfont{CFL-W} start with each client in one cluster, and recursively split clusters over rounds and over time based on the intra-cluster similarity of their local updates. We observe that the \algfont{CFL-W} variant is the better-performing of the two on each dataset except MNIST-4 (which is also the only dataset where \algfont{Oblivious} outperforms \algfont{Window}), and is a consequence of the passive drift adaptation of its sliding window which forgets older data. The performance of \algfont{CFL-W} is relatively high on SINE-2 and CIRCLE-2. As an example, the clustering learned on SINE-2 is shown in Figure \ref{fig:cfl}. We observe that, for the first 6 time steps, it correctly distinguishes the two concepts by using distinct models. The disadvantage of the clustering of \algfont{CFL-W} is that it creates excess models for the same concept and does not take full advantage of collaborative training. At time 5, it is limited to splitting its cluster for model 0 when the green concept occurs, but cannot merge the drifted clients to the existing cluster created for the green concept at the previous time step. This limitation of only being able to subdivide existing clusters, but not merge clusters or re-assign clients to existing clusters results in poor performance on more complex drifts.

For \algfont{IFCA}, \algfont{IFCA-W}, and \algfont{IFCA(T)}, the clustering is pre-initialized with a random model for each concept that can occur over time for each dataset. In general, we observe that this is not a reliable method for reacting to drift. All the \algfont{IFCA} variants perform well under the sharp label-swap drift of SINE-2. When the new concept occurs, the drifted clients cluster to the second model, and the learned clustering matches the ground-truth. On CIRCLE-2, we found that \algfont{IFCA} and \algfont{IFCA(T)} learned the correct clustering in 2 out of 5 trials, and otherwise used only a single model in the other 3 trials. \algfont{IFCA-W} learned the correct clustering in 1 out of 5 trials. (Note the high standard deviation in Table \ref{table:accuracy-extra}.) Across the SEA and MNIST datasets, none of the three algorithms ever used more than a single model (with one exception---on SEA-4, in 1 out of 5 trials, \algfont{IFCA-W} used a distinct model for the yellow concept). For the SEA and MNIST datasets, we observe that the \algfont{IFCA} and \algfont{IFCA(T)} degrade to the \algfont{Oblivious} algorithm, and that \algfont{IFCA-W} degrades to the \algfont{Window} algorithm. Note that despite the degenerate clustering to a single model, matching the \algfont{Window} algorithm, \algfont{IFCA-W} achieves the highest accuracy in Table \ref{table:accuracy-extra} on SEA-2 (but within the standard deviation of \algfont{Window}) due to randomness in the model initialization. In general, our experimental protocol fixes the same random seed for each trial across all all algorithms and where all created models within an algorithm use the same initialization. The \algfont{IFCA} variants are an exception to the rule because its initialization requires \emph{distinct} random models. On the FMoW dataset, we observe again that random initialization can sometimes address drift, but unreliably: in 1 out of 5 trials each for all \algfont{IFCA} variants, a separate model is used for the Africa region at later time steps. (However, the \algfont{IFCA} variants are among the worst performing in our evaluation because their random initialization precludes the pre-trained ImageNet initialization we use for other algorithms.) The authors of the original paper on \algfont{IFCA} note that the accuracy of the clustering is sensitive to the initialization of the models, and propose random restarts to address this issue, but restarts do not translate well to the time-varying setting we study. In our work, \mmacc and \name address the initialization problem by using drift detection to deal with new concepts as they occur and to cultivate new clusters.

For \algfont{\mmacc-W} and \algfont{\name-W}, restricting to a window has minimal impact on the accuracy for the SEA dataset.
There is a significant loss of accuracy for the MNIST dataset relative to the non-windowed versions, but note that the same significant loss occurs when going from \algfont{Oracle} to \algfont{Oracle-W}, so this loss is a result of windowing, not specific to our algorithm.  Indeed, the accuracy of \algfont{\name-W} is quite close to \algfont{Oracle-W}.

\paragraph{The communication-efficient \algfont{FedDrift-C}.} As noted in \S\ref{sec:algo-creation}, one of the drawbacks of \name is that it can create more models $M$ compared to \mmacc, adding to the communication cost  of sending $O(MP)$ models. The goal is to only use a number of global models close or equal to the number of distinct concepts, and while \name can hierarchically merge created models of the same concept, \name can observe temporary spikes in the number of global models. To mitigate this cost, we evaluate \algfont{FedDrift-C}, which differs from \name in that, at each time after drift occurs, only one random client that drifted contributes its local model as a global model. In the case that multiple new concepts occur at a time, only one of the new concepts will be learned immediately, but clients that are still at an unlearned concept are eligible to detect drift again at the following time step and get another chance to contribute its local model. Meanwhile, while a concept goes unlearned globally, drifted clients do not contribute to any of the global models. 

For the 4 concepts in MNIST-4, we observed that \name learned a total of 7 global models (later merged down to 4) as shown in Figure \ref{fig:mnist-clustering} in \S\ref{sec:expt}. \algfont{FedDrift-C} more efficiently maintained a maximum of 4 global models across all time, at a penalty of 0.92\% accuracy due to the delayed learning of one of the two simultaneously arising concepts. Meanwhile, \mmacc suffers a larger 3.11\% penalty after it incorrectly merged the two simultaneous concepts, as shown in Figure \ref{fig:mmacc-mnist4}---model 1 is initially trained over the green and yellow concepts, and while the clients at the green concept later abandon model 1 and eventually learn a separate model 2, the green concept training data still poison both model 0 and model 1.

\begin{figure}[h!]
\centering
\includegraphics[width=0.45\linewidth]{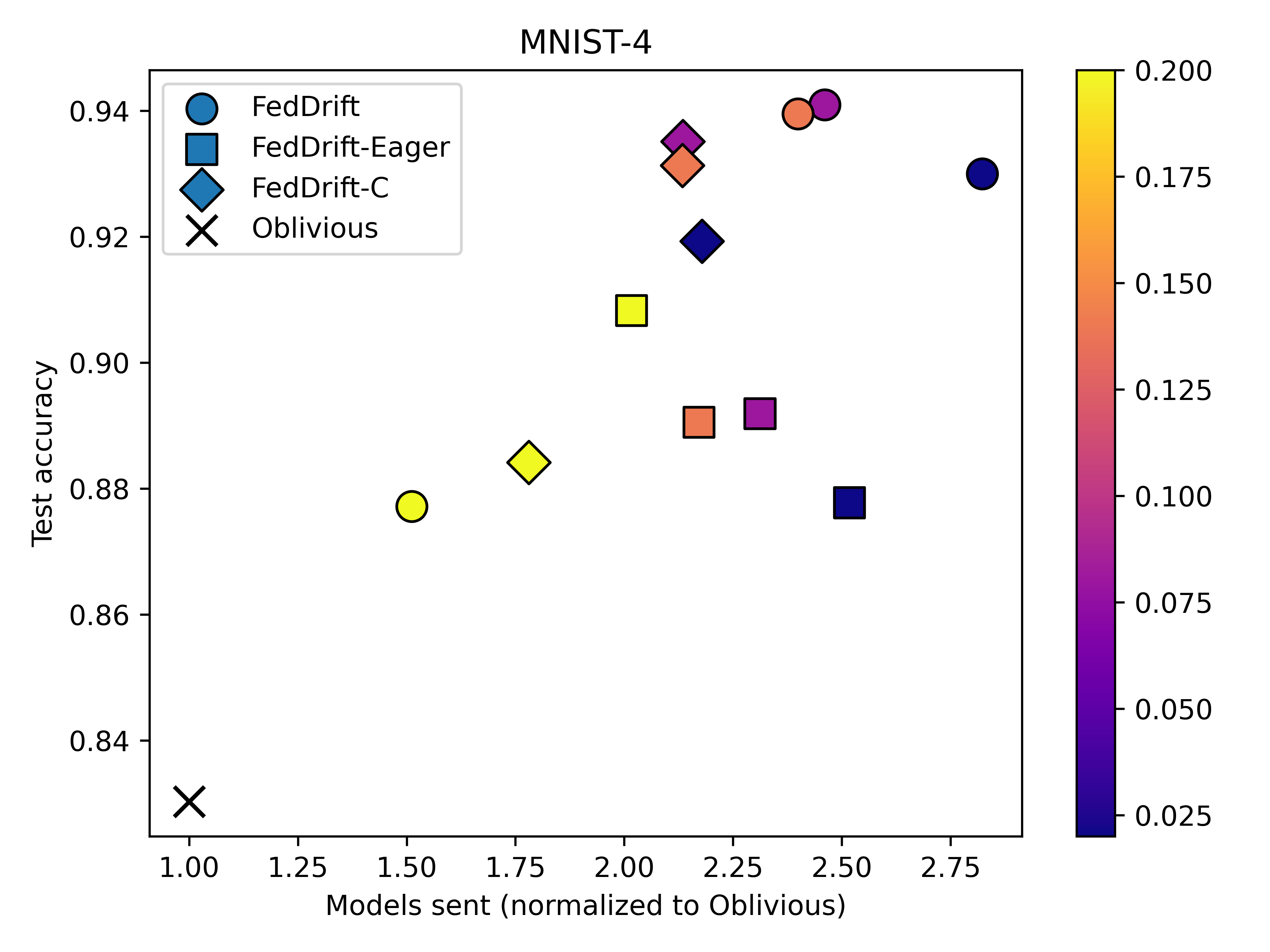}
\caption{The accuracy-communication trade-off on MNIST-4 for \mmacc, \name, and \algfont{FedDrift-C}. Each algorithm is evaluated under various selections of the splitting/merging threshold $\delta$ between 0.02 and 0.20, indicated by color. The vertical axis is the average test accuracy across clients and time, omitting drifts. (1 trial)}
\label{fig:accuracy-communication}
\end{figure}

We quantify this accuracy-communication trade-off in Figure \ref{fig:accuracy-communication} where we show the average test accuracy and total number of models sent by \mmacc, \name, and \algfont{FedDrift-C} under various selections of the drift detection threshold $\delta$. Increasing the value of $\delta$ restricts cluster splitting (increases false negative detections) and promotes cluster merging, which reduces the number of models and concepts learned (at $\delta = 1$, each algorithm is identical to \algfont{Oblivious}). Empirically, we confirm that choosing larger settings of $\delta$ can trade-off accuracy for efficiency. (Choosing $\delta$ too small for \name can also negatively affect accuracy due to increased false positive detections, but to a lesser degree because the hierarchical clustering of \name can correct some false positives---see below on Impact of False Positives.) We observe that, generally, using \algfont{FedDrift-C} over \name preserves most of the accuracy improvement over \algfont{Oblivious} while saving communication---with one exception at the largest $\delta=0.20$ where both algorithms are susceptible to false merging, but \name has more total models added to make the mistake of merging two concepts that \algfont{FedDrift-C} avoids. We also observe that the Pareto front is mostly configurations of \name and \algfont{FedDrift-C} over \mmacc. Finally, we observe that all variants of \name are more efficient than ensemble algorithms---relative to \algfont{Oblivious}, \name variants send 2--3x models compared to AUE which sends 5x---because for ensembles, clients contribute to every model at each communication round, compared to \name where clients contribute only to the clusters they belong to (the broadcast of all models for clustering in \name is only once per time step).

\paragraph{Random Drift Patterns.} Throughout this paper, we have considered the 4-concept drift pattern in Figure \ref{fig:4concepts} in \S\ref{sec:motivation} as a specific concrete example in order to depict the challenges in distributed concept drift, motivate the design of \name, and discuss the experimental performance by comparing the learned clustering matrix to the ground-truth. To examine the performance more generally, we consider a family of datasets MNIST-R with random concept changes. Using the same four concepts as in MNIST-4, MNIST-R is generated with all clients at the first concept to start, and then each client independently randomly observes one of the four concepts every two time steps (as opposed to every time step which is not possible to adapt to).  Across 5 random seeds, the average accuracy is shown in Table \ref{table:mnist-random} (and in Table \ref{table:mnist-random-drift} for all time including drifts).
We generally observe the same relative performances of each algorithm as on the previously specified MNIST-4 drift. The performance of \name is close to that of \algfont{Oracle}, \algfont{FedDrift-C} is close behind, \mmacc is lower given that it is likely to have multiple new concepts occurring simultaneously in MNIST-R, and then all prior baselines follow.

\begin{table}[t]
    \begin{minipage}{.5\linewidth}
    \caption{Accuracy (\%) on MNIST-R, omitting drifts}
    \label{table:mnist-random}
    \begin{center}
    \begin{small}
    \begin{tabular}{lc}
    \toprule
     & MNIST-R \\
    \midrule
    \algfont{Oblivious} &  	85.12	$\pm$	1.37	\\
    \algfont{DriftSurf} &	85.03	$\pm$	1.36	\\
    \algfont{KUE} &	81.56	$\pm$	1.90	\\
    \algfont{AUE} &	83.87	$\pm$	1.64	\\
    \algfont{AUE-PC} &	83.67	$\pm$	1.66	\\
    \algfont{Window} &	82.37	$\pm$	1.94	\\
    \algfont{Window-2} &	83.65	$\pm$	1.83	\\
    \algfont{Weighted-Linear} &	84.87	$\pm$	1.34	\\
    \algfont{Weighted-Exp}   &	84.60	$\pm$	1.44	\\
    \algfont{Adaptive-FedAvg} &	83.17	$\pm$	1.51	\\
    \algfont{CFL}            &	84.20	$\pm$	1.54	\\
    \algfont{CFL-W}          &	82.24	$\pm$	1.77	\\
    \algfont{IFCA(T)} &	84.50	$\pm$	1.21	\\
    \algfont{IFCA} &	84.39	$\pm$	1.45	\\
    \algfont{IFCA-W} &	85.93	$\pm$	3.35	\\
    \midrule				
    \mmacc &	89.85	$\pm$	1.49	\\
    \name &	\textbf{94.06	$\pm$	0.38}	\\
    \algfont{FedDrift-C} &	92.76	$\pm$	0.56	\\
    \algfont{FedDrift-Eager-W} &	86.60	$\pm$	2.27	\\
    \algfont{FedDrift-W} &	90.83	$\pm$	0.17	\\
    \midrule				
    \algfont{Oracle} &	95.03	$\pm$	0.15	\\
    \algfont{Oracle-W} &	91.66	$\pm$	0.31	\\
    \bottomrule
    \end{tabular}
    \end{small}
    \end{center}
    \end{minipage}
    \begin{minipage}{.5\linewidth}
    \caption{Accuracy (\%) on MNIST-R, including drifts}
    \label{table:mnist-random-drift}
    \begin{center}
    \begin{small}
    \begin{tabular}{lc}
    \toprule
     & MNIST-R \\
    \midrule
     \algfont{Oblivious} &  	83.92	$\pm$	1.23	\\
    \algfont{DriftSurf} &	83.83	$\pm$	1.21	\\
    \algfont{KUE} &	79.77	$\pm$	2.03	\\
    \algfont{AUE} &	81.96	$\pm$	1.03	\\
    \algfont{AUE-PC} &	81.52	$\pm$	1.43	\\
    \algfont{Window} &	80.11	$\pm$	1.45	\\
    \algfont{Window-2} &	81.30	$\pm$	1.58	\\
    \algfont{Weighted-Linear} &	83.64	$\pm$	1.21	\\
    \algfont{Weighted-Exp}   &	83.39	$\pm$	1.29	\\
    \algfont{Adaptive-FedAvg} &	81.41	$\pm$	1.24	\\
    \algfont{CFL}            &	83.05	$\pm$	1.37	\\
    \algfont{CFL-W}          &	80.58	$\pm$	1.94	\\
    \algfont{IFCA(T)} &	83.31	$\pm$	1.11	\\
    \algfont{IFCA} &	83.29	$\pm$	1.29	\\
    \algfont{IFCA-W} &	81.65	$\pm$	0.67	\\
  \midrule				
    \mmacc &	85.26	$\pm$	0.81	\\
    \name &	\textbf{86.77	$\pm$	0.76}	\\
    \algfont{FedDrift-C} &	86.65	$\pm$	0.94	\\
    \algfont{FedDrift-Eager-W} &	81.74	$\pm$	1.60	\\
    \algfont{FedDrift-W} &	83.70	$\pm$	0.80	\\
    \midrule				
    \algfont{Oracle} &	87.32	$\pm$	0.86	\\
    \algfont{Oracle-W} &	84.29	$\pm$	0.89	\\
    \bottomrule
    \end{tabular}
    \end{small}
    \end{center}
    \end{minipage}
\end{table}

\paragraph{Adaptation under Label Shift.}
In this work, we focus on the general case of concept drift as opposed to special cases like covariate shift or label shift. All of the synthetic drift datasets studied above involve a change in the decision boundary. But our solutions are also applicable under specific cases of drift. The real drift in the FMoW dataset is an example of label shift (defined in \S\ref{sec:setup}). Here we consider another label shift dataset, MNIST-L-4, in which the drift is synthetically generated to follow the same distributed drift pattern as MNIST-4 in the ground-truth clustering, but differs in the concept definitions so that classes are incrementally introduced over time: concept A is only over digits 0/1, concept B is only over digits 2/3, concept C is only over digits 4/5, and concept D is only over digits 6/7. 

Table \ref{table:mnist-label} shows the average accuracy omitting time steps of drift, and Table \ref{table:mnist-label-drift} shows the average accuracy over all time including drift times. Unlike all previous datasets (including FMoW with label shift), we observe a significant difference between the two tables. On the metric omitting drifts, \mmacc and \name attain 99\% accuracy comparable to \algfont{Oracle}, then the \algfont{IFCA} variants attain similarly high accuracy compared to the remaining baselines. However, on the metric including drifts, we observe \algfont{Adaptive-FedAvg} performs best, and even the \algfont{Oblivious} algorithm outperforms the multiple-model algorithms like \algfont{FedDrift}, \algfont{IFCA}, and \algfont{Oracle}. The reason is that for the concepts in MNIST-L-4, it is possible for a single model to fit multiple concepts (different labels) accurately as long as a concept was previously seen at another client. On the other hand, the multi-model approach suffers from poor performance at the time of drift as the newly created model has not seen the labels in the test data. Under the 4-concept drift pattern, the single model learned by \algfont{Adaptive-FedAvg} has high test accuracy on concepts B (green) and C (yellow) after the first occurrence in the system, while \algfont{Oracle} has low accuracy at the time of drift by employing a specialized model trained solely on concept A (blue).

\begin{table}[t]
    \begin{minipage}{.5\linewidth}
    \caption{Accuracy (\%) on MNIST-L-4, omitting drifts}
    \label{table:mnist-label}
    \begin{center}
    \begin{small}
    \begin{tabular}{lc}
    \toprule
     & MNIST-L-4 \\
    \midrule
\algfont{Oblivious}	&	90.07	$\pm$	0.23	\\
\algfont{DriftSurf}	&	91.79	$\pm$	0.64	\\
\algfont{KUE}	&	90.68	$\pm$	1.03	\\
\algfont{AUE}	&	92.76	$\pm$	1.29	\\
\algfont{AUE-PC}	&	92.08	$\pm$	1.40	\\
\algfont{Window}	&	87.28	$\pm$	1.26	\\
\algfont{Window-2}	&	90.10	$\pm$	0.33	\\
\algfont{Weighted-Linear}	&	91.13	$\pm$	0.56	\\
\algfont{Weighted-Exp}	&	91.87	$\pm$	0.31	\\
\algfont{Adaptive-FedAvg}	&	96.74	$\pm$	0.04	\\
\algfont{CFL} 	&	87.92	$\pm$	1.52	\\
\algfont{CFL-W}	&	92.00	$\pm$	1.25	\\
\algfont{IFCA(T)}	&	98.86	$\pm$	1.01	\\
\algfont{IFCA}	&	99.48	$\pm$	0.34	\\
\algfont{IFCA-W}	&	99.02	$\pm$	0.62	\\
\midrule					
\mmacc	&	99.17	$\pm$	0.09	\\
\name 	&	\textbf{99.56	$\pm$	0.01}	\\
\algfont{FedDrift-C}	&	97.85	$\pm$	0.86	\\
\algfont{FedDrift-Eager-W}	&	98.16	$\pm$	0.80	\\
\algfont{FedDrift-W}	&	99.20	$\pm$	0.02	\\
\midrule					
\algfont{Oracle}	&	99.65	$\pm$	0.01	\\
\algfont{Oracle-W}	&	99.29	$\pm$	0.01	\\

    \bottomrule
    \end{tabular}
    \end{small}
    \end{center}
    \end{minipage}
    \begin{minipage}{.5\linewidth}
    \caption{Accuracy (\%) on MNIST-L-4, including drifts}
    \label{table:mnist-label-drift}
    \begin{center}
    \begin{small}
    \begin{tabular}{lc}
    \toprule
     & MNIST-L-4 \\
    \midrule
\algfont{Oblivious}	&	81.60	$\pm$	0.29	\\
\algfont{DriftSurf}	&	83.48	$\pm$	1.24	\\
\algfont{KUE}	&	81.44	$\pm$	1.07	\\
\algfont{AUE}	&	83.58	$\pm$	1.22	\\
\algfont{AUE-PC}	&	83.08	$\pm$	1.23	\\
\algfont{Window}	&	77.14	$\pm$	1.30	\\
\algfont{Window-2}	&	81.34	$\pm$	0.22	\\
\algfont{Weighted-Linear}	&	82.97	$\pm$	0.66	\\
\algfont{Weighted-Exp}	&	83.89	$\pm$	0.33	\\
\algfont{Adaptive-FedAvg}	&	\textbf{89.44	$\pm$	0.03}	\\
\algfont{CFL} 	&	77.08	$\pm$	0.81	\\
\algfont{CFL-W}	&	81.67	$\pm$	1.59	\\
\algfont{IFCA(T)}	&	71.99	$\pm$	2.20	\\
\algfont{IFCA}	&	71.98	$\pm$	2.79	\\
\algfont{IFCA-W}	&	71.64	$\pm$	2.54	\\
\midrule					
\mmacc	&	75.15	$\pm$	1.30	\\
\name 	&	70.69	$\pm$	0.01	\\
\algfont{FedDrift-C}	&	69.79	$\pm$	0.60	\\
\algfont{FedDrift-Eager-W}	&	74.94	$\pm$	1.23	\\
\algfont{FedDrift-W}	&	70.43	$\pm$	0.01	\\
\midrule					
\algfont{Oracle}	&	70.75	$\pm$	0.01	\\
\algfont{Oracle-W}	&	70.49	$\pm$	0.01	\\
    \bottomrule
    \end{tabular}
    \end{small}
    \end{center}
    \end{minipage}
\end{table}

\begin{figure}[h]
\centering
    \begin{subfigure}[t]{0.30\linewidth}
        \includegraphics[width=\linewidth]{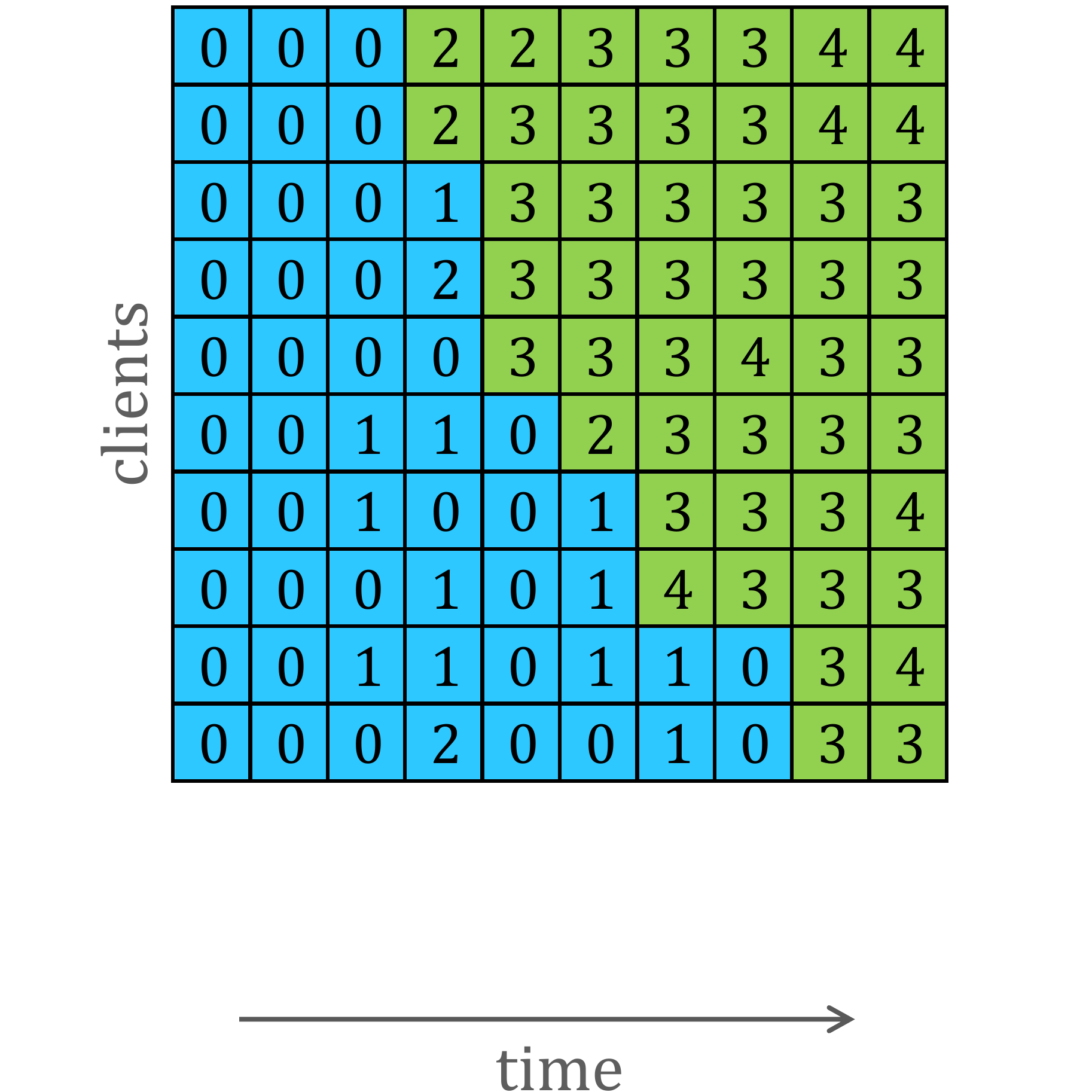}
        \caption{\mmacc}
        \label{fig:mmacc-sine2}
    \end{subfigure}
   \hspace{0.1in}
    \begin{subfigure}[t]{0.30\linewidth}
        \includegraphics[width=\linewidth]{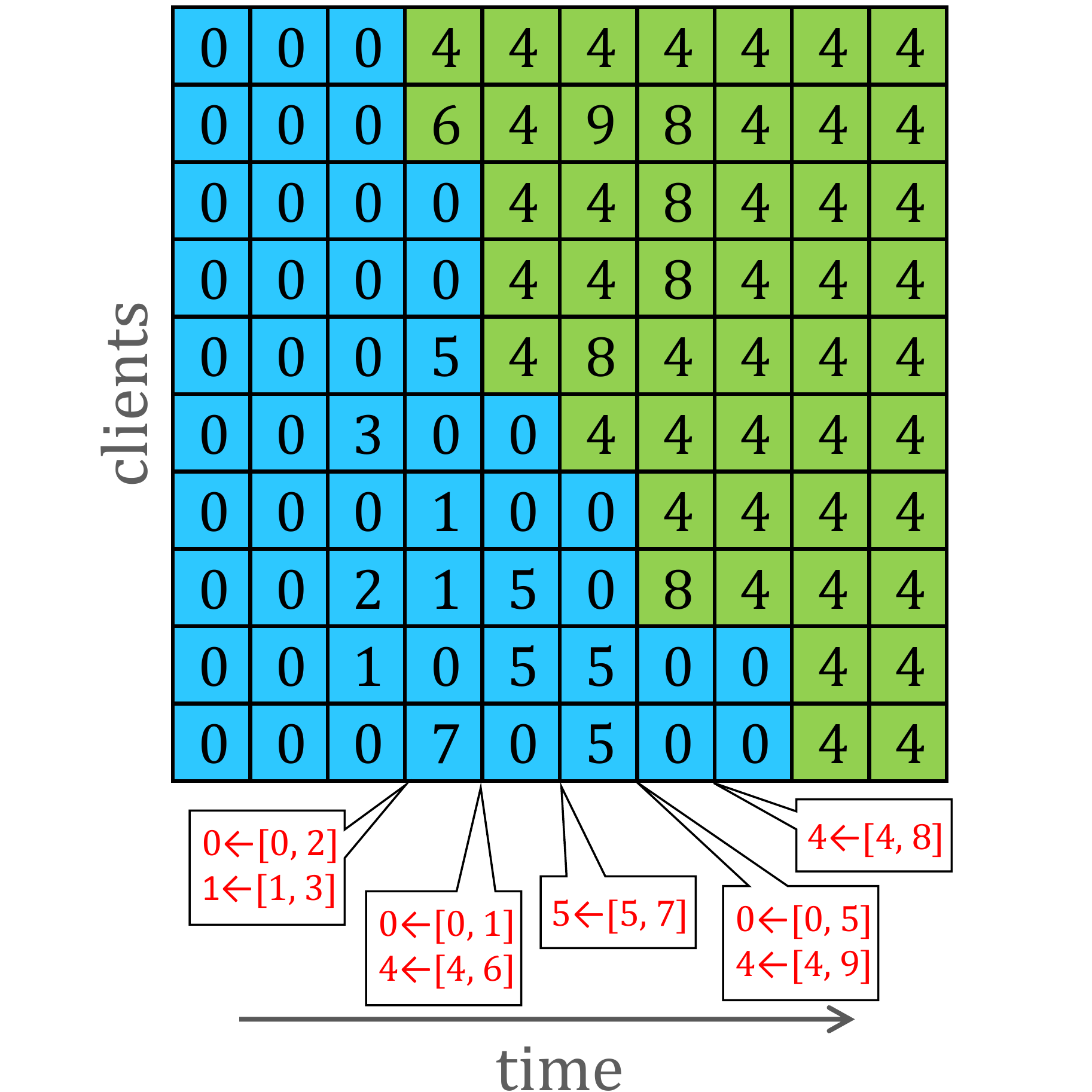}
        \caption{\name}
        \label{fig:feddrift-sine2}
    \end{subfigure}
\caption{The clustering learned on SINE-2 when $\delta=0.01$. Each cell indicates the model ID at each client and time step, and the background color indicates the ground-truth concept.}    
\label{fig:sine-clustering}
\end{figure}

\paragraph{Impact of False Positives.}
To demonstrate the application of the hierarchical clustering in \name, in \S\ref{sec:expt} we discussed the example of the learned clustering for MNIST-4 in Figure \ref{fig:mnist-clustering}. Here in Figure \ref{fig:sine-clustering} we present another example on SINE-2 at a small $\delta=0.01$ (corresponding to more aggressive detection) to demonstrate an example of how hierarchical clustering can be beneficial even in the case of a 2-concept drift in mitigating false positives. At time 3, in both \mmacc and \name there are three false positives, where in \mmacc, the new model 1 is retained but its underlying data forgotten, while in \name, although initially 3 redundant models are created, they are all merged back with model 0 within 2 time steps, averaging their parameters and reincorporating their clustered data. The advantage of hierarchical clustering is also evident at time 4 when 2 false positives and 2 true positives occur together. In \mmacc, one new model is created for all the clients, but this new model is ``poisoned'' by contributions from the blue concept and does not work well at time 5, resulting in another drift detection to create model 3 (and forgetting about the data associated with model 2). \name, on the other hand, creates models solely trained over either the blue and green concepts, and eventually merges all models of an identical concept, recovering all of the data. While the false positive mitigation demonstrated in this example is not a significant contributor to the observed higher accuracy of \name in our evaluation because we use higher $\delta$ values as noted in Appendix \ref{sec:appendix-expt-setup}, it is relevant when there is greater uncertainty in selecting the threshold hyperparameter.

\begin{figure}[h!]

\begin{center}
    \begin{subfigure}[t]{0.4\textwidth}
        \includegraphics[trim=1cm 8cm 0.5cm 7cm, width=\columnwidth]{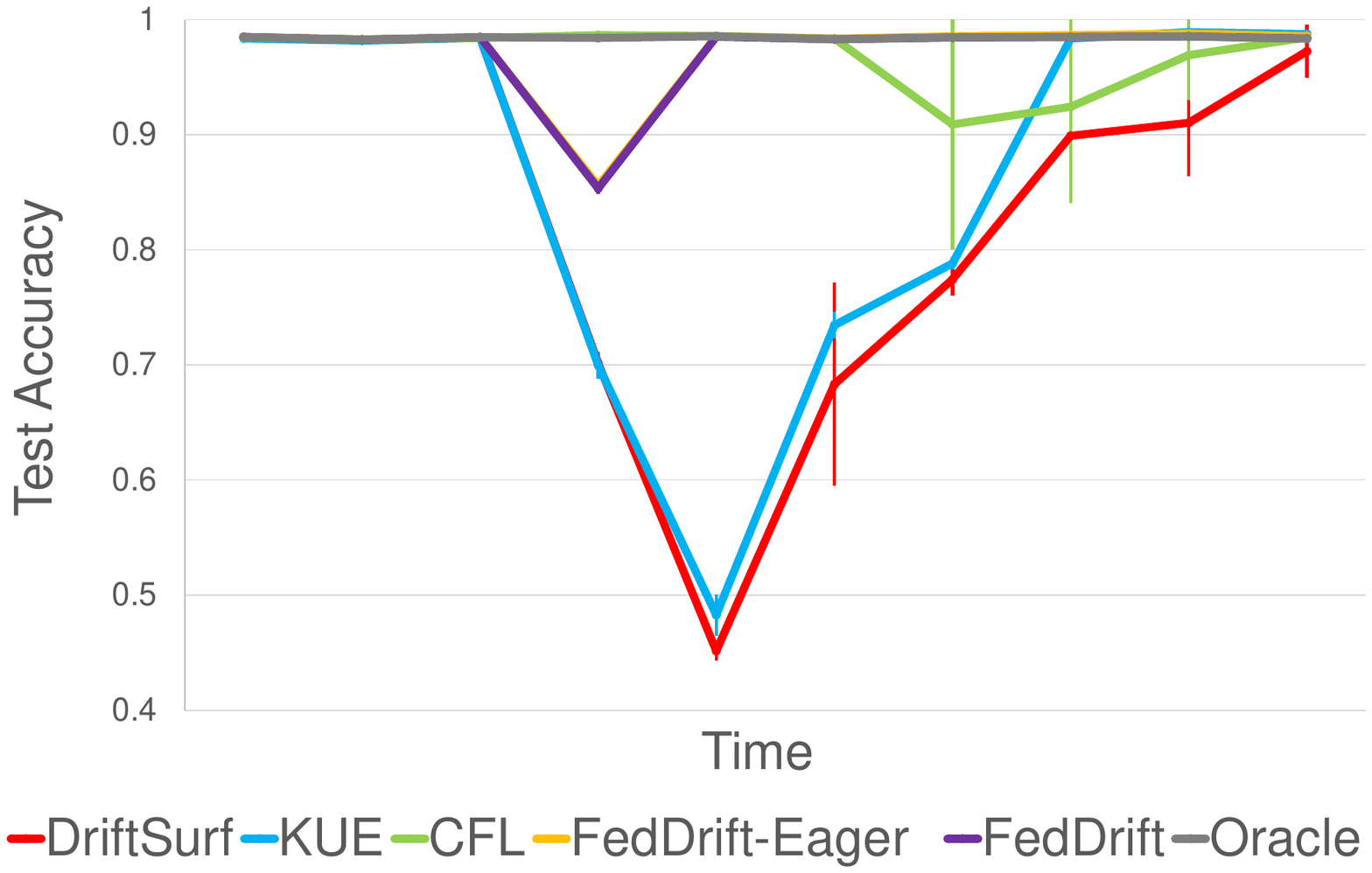}
        \caption{SINE-2}
        \label{fig:sine2}
    \end{subfigure}\hspace{0.3in}
    \begin{subfigure}[t]{0.4\textwidth}
        \includegraphics[trim=1cm 8cm 0.5cm 7cm, width=\columnwidth]{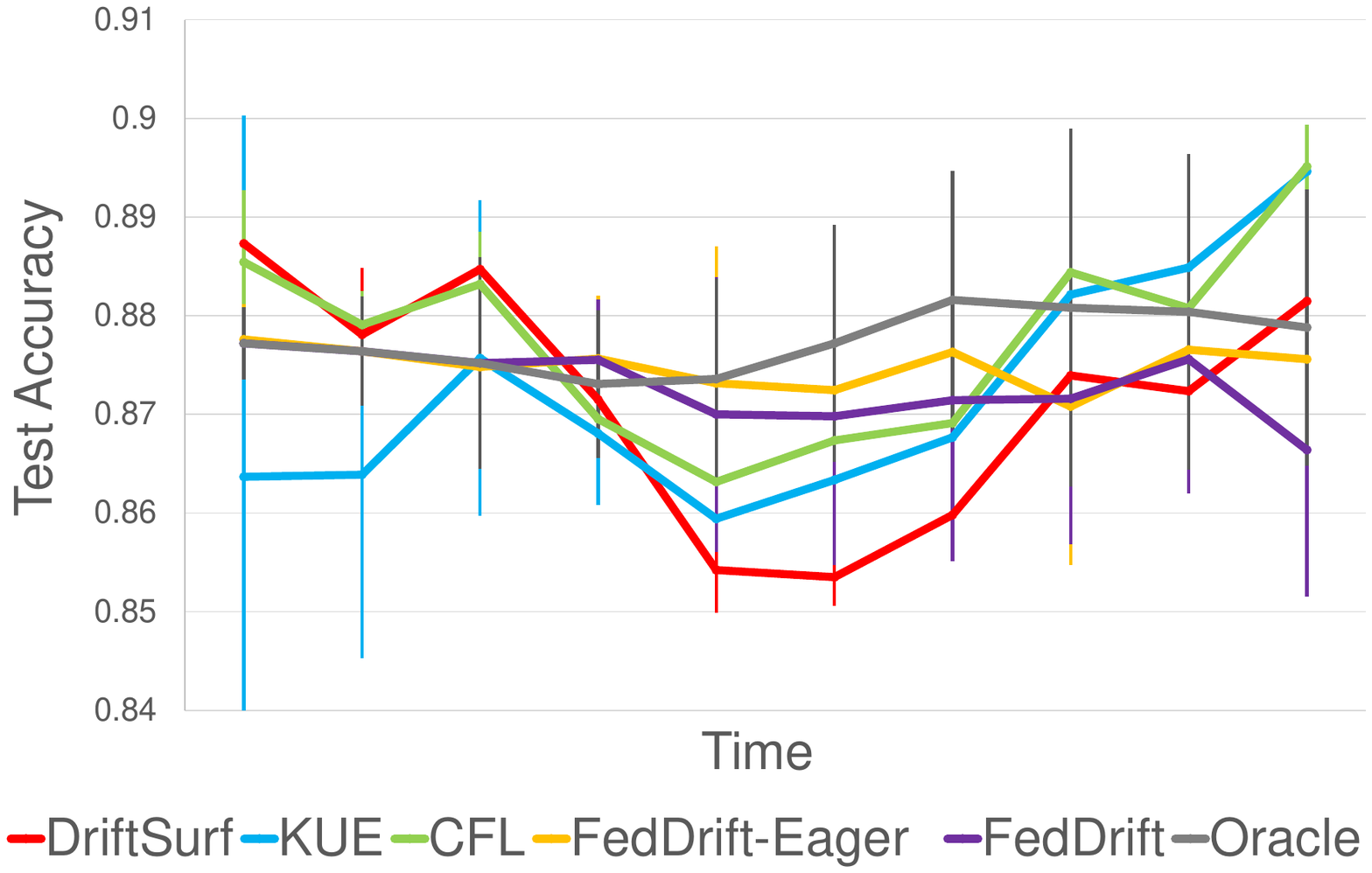}
        \caption{SEA-2}
        \label{fig:sea2}
    \end{subfigure}\hfill

\vspace{-0.1in}

    \begin{subfigure}[t]{0.4\textwidth}
        \includegraphics[trim=1cm 8cm 0.5cm 7cm, width=\columnwidth]{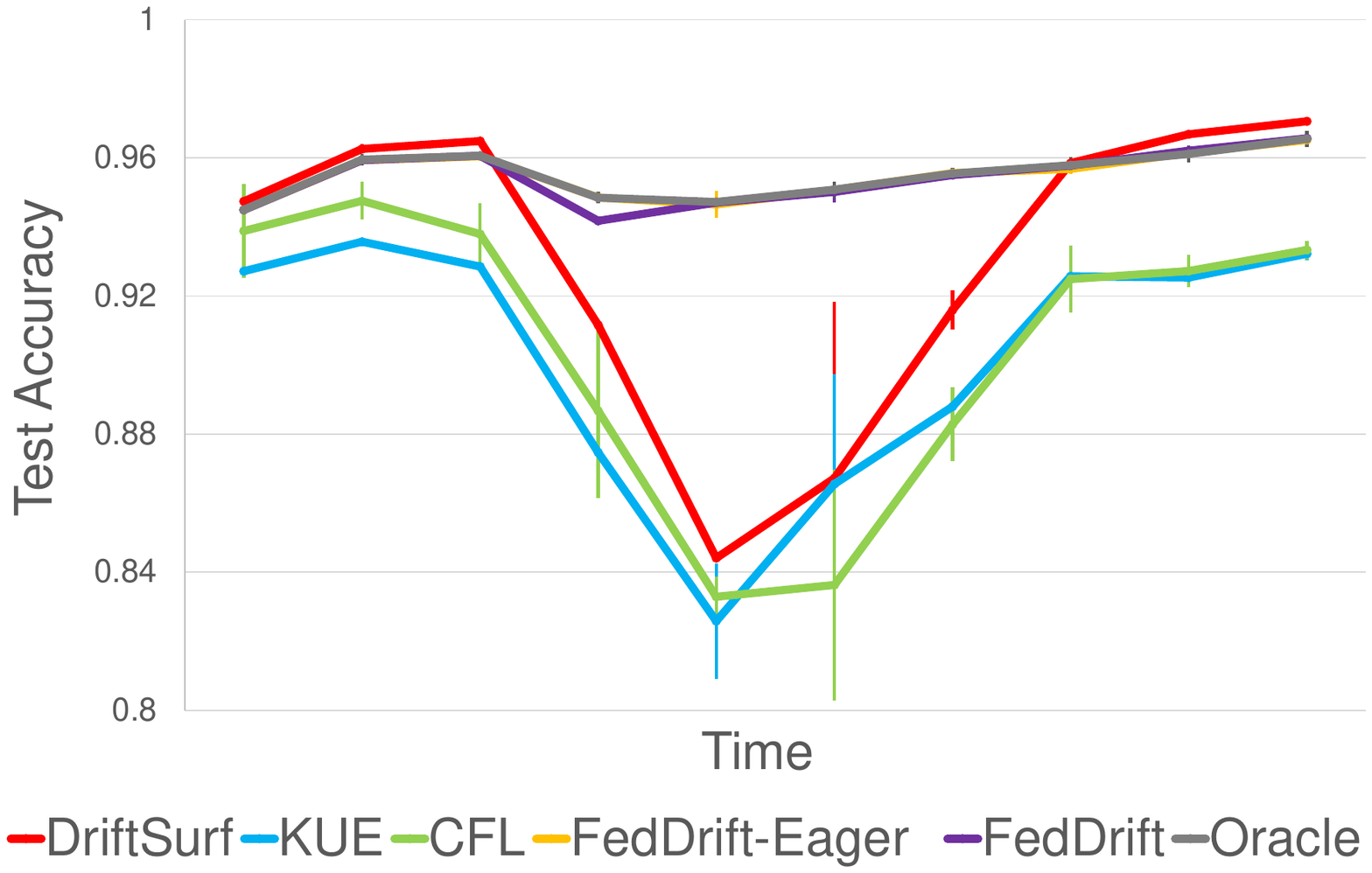}
        \caption{MNIST-2}
        \label{fig:mnist2}
    \end{subfigure}\hspace{0.3in}
    \begin{subfigure}[t]{0.4\textwidth}
        \includegraphics[trim=1cm 8cm 0.5cm 7cm, width=\columnwidth]{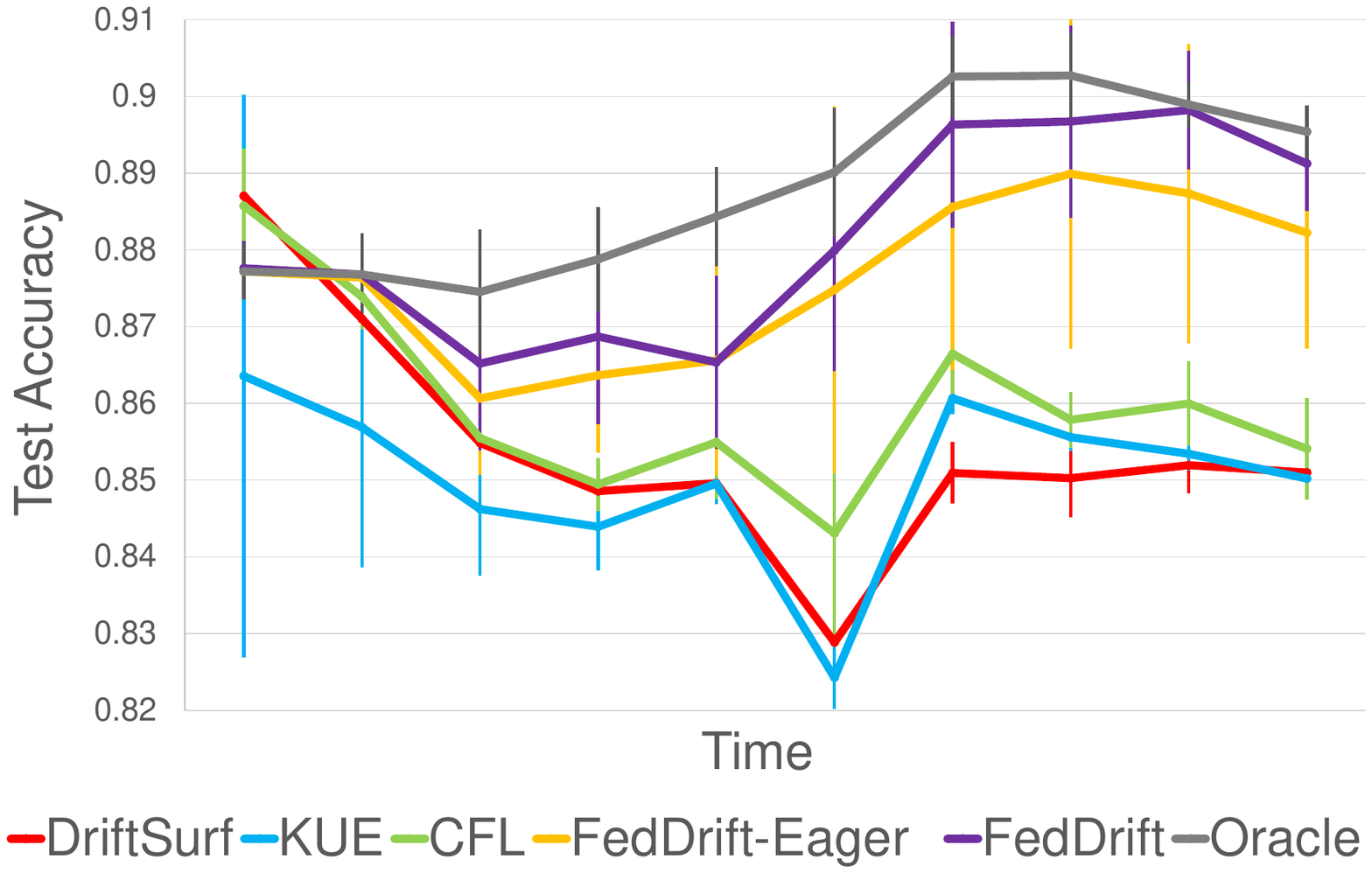}
        \caption{SEA-4}
        \label{fig:sea4}
    \end{subfigure}\hfill    

\vspace{-0.1in}
    
    \begin{subfigure}[t]{0.4\textwidth}
        \includegraphics[trim=1cm 8cm 0.5cm 7cm, width=\columnwidth]{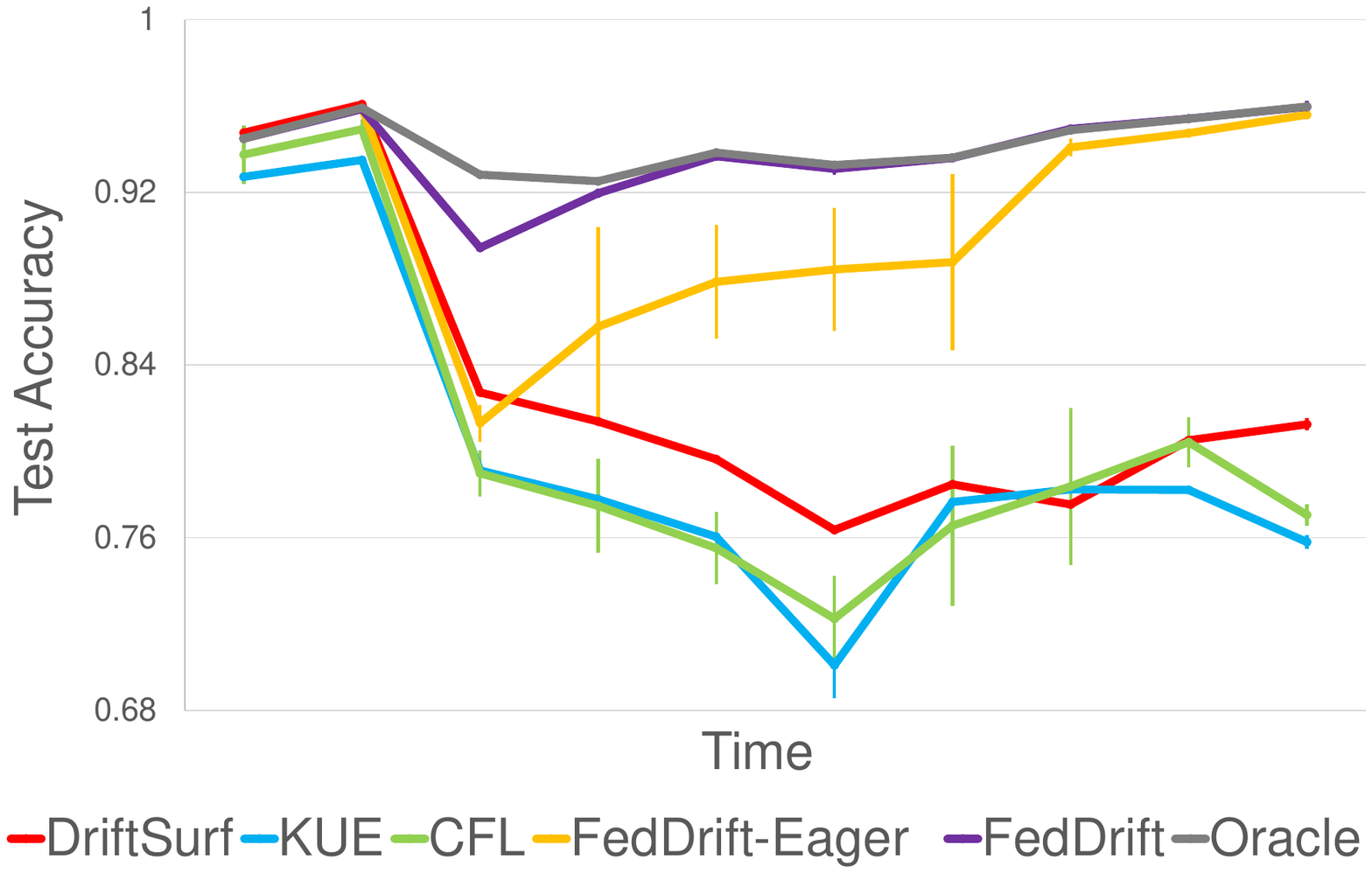}
        \caption{MNIST-4}
        \label{fig:mnist4}
    \end{subfigure}\hspace{0.3in}
    \begin{subfigure}[t]{0.4\textwidth}
        \includegraphics[trim=1cm 8cm 0.5cm 7cm, width=\columnwidth]{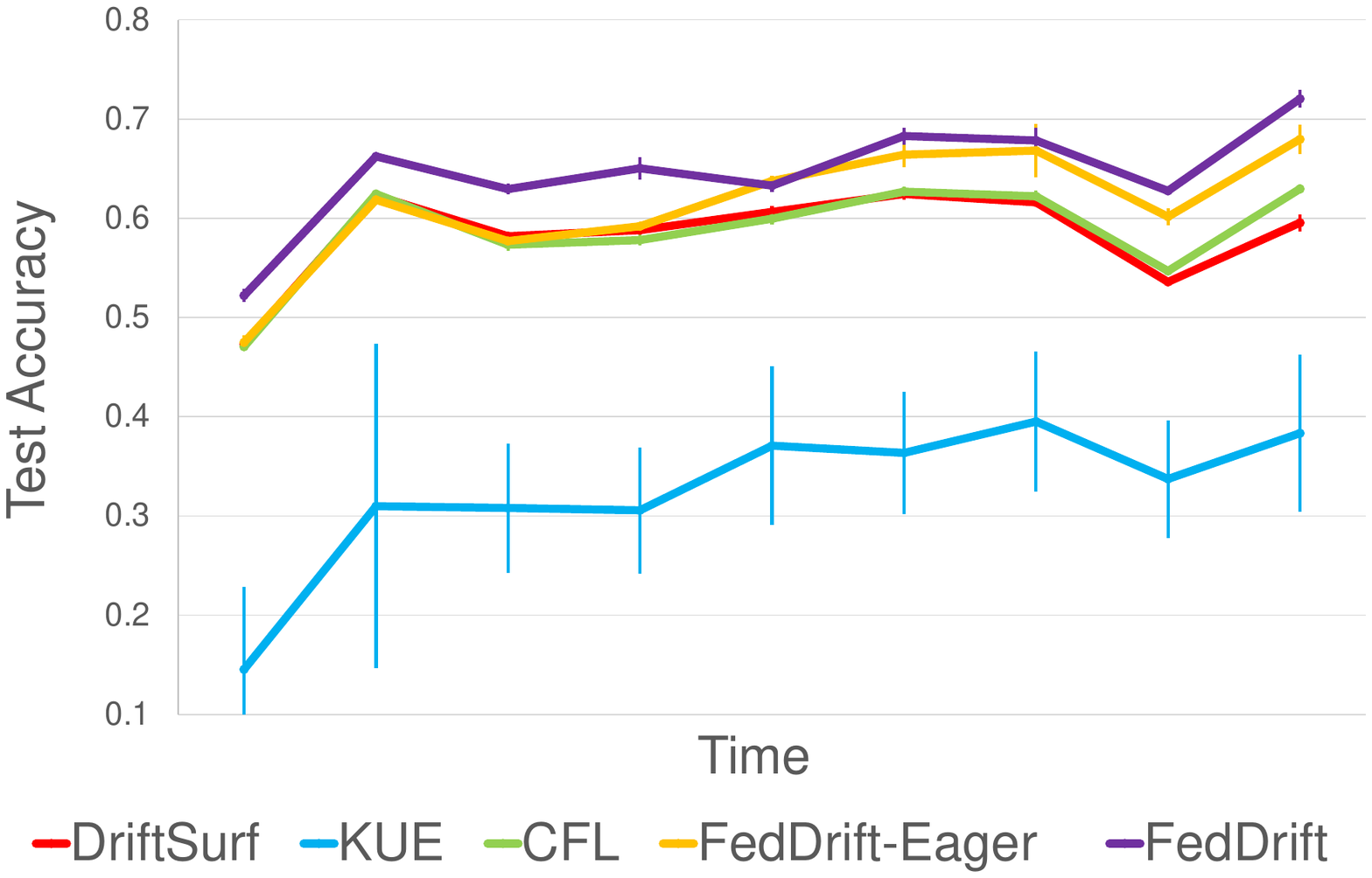}
        \caption{FMoW}
        \label{fig:fmow}
    \end{subfigure}\hfill    
\end{center}
\vspace{-0.1in}
\caption{Test accuracy of selected algorithms at each time on SINE-2, SEA-2, MNIST-2, SEA-4, MNIST-4, and FMoW. Vertical lines represent standard deviations.}
\label{fig:time-extras}
\end{figure}

\paragraph{Test Accuracy Over Time.}
Finally, in Figure \ref{fig:time-extras}, we include plots that we omit from the main paper due to space constraints. The figure shows the accuracy over time for \mmacc, \name, and selected baselines representing drift detection, ensembles, and clustered FL, supplementing Figure \ref{fig:time} in \S\ref{sec:expt}. (Note the varying scales of the y-axes.) We observe the same general trends: (i) the centralized drift adaptation algorithms suffer in performance, particularly during the transition period when no one model works well across all clients; (ii) CFL can react to the drift early on SINE-2 as with CIRCLE-2 before, but its performance degrades with excessive further splits; (iii) for the 4-concept drift in SEA-4 and MNIST-4 centralized baselines and CFL never recover in performance with multiple concepts present; and (iv) on SEA-4 and MNIST-4, \name is close to \algfont{Oracle} except for a gap at time 3 when it uses local models prior to merging, while \mmacc lags behind \name when it creates a single model for the 2 simultaneously arising concepts but can slowly recover with further detections.

\end{document}